\documentclass[10pt,journal,compsoc]{IEEEtran}

\usepackage{epsfig}
\usepackage{graphicx}
\usepackage{amsmath}
\usepackage{amssymb}

\usepackage{float}
\usepackage{booktabs}
\usepackage{caption}
\usepackage{subcaption}
\usepackage{multirow}
\usepackage{xcolor}
\usepackage{cite}
\usepackage[utf8]{inputenc}
\usepackage[vlined,ruled]{algorithm2e}
\usepackage{cuted}
\usepackage[pagebackref=true,breaklinks=true,colorlinks,bookmarks=false]{hyperref}
\usepackage{url}
\usepackage{ragged2e}

\newcommand{\ba}{\mathbf{a}}\newcommand{\bA}{\mathbf{A}}

\newcommand{\bF}{\mathbf{F}} %

\newcommand{\bK}{\mathbf{K}}

\newcommand{\bM}{\mathbf{M}}

\newcommand{\bO}{\mathbf{O}}
\newcommand{\bP}{\mathbf{P}}
\newcommand{\bQ}{\mathbf{Q}}
\newcommand{\bR}{\mathbf{R}}

\newcommand{\bV}{\mathbf{V}}
\newcommand{\bw}{\mathbf{w}}

\newcommand{\nE}{\mathbb{E}}

\newcommand{\nI}{\mathbb{I}}

\newcommand{\nR}{\mathbb{R}}

\newcommand{\cD}{\mathcal{D}}

\newcommand{\cG}{\mathcal{G}}

\newcommand{\cL}{\mathcal{L}}

\newcommand{\cW}{\mathcal{W}}
\newcommand{\cX}{\mathcal{X}}

\newcommand{\figref}[1]{Fig.~\ref{#1}}
\newcommand{\secref}[1]{Section~\ref{#1}}

\newcommand{\tabref}[1]{Table~\ref{#1}}

\DeclareMathOperator*{\argmin}{argmin~}

\makeatletter
\DeclareRobustCommand\onedot{\futurelet\@let@token\@onedot}
\def\@onedot{\ifx\@let@token.\else.\null\fi\xspace}
\def\eg{e.g\onedot} 
\def\ie{i.e\onedot}

\makeatother

\newcommand{\boldparagraph}[1]{\vspace{0.2cm}\noindent{\bf #1:}}

\definecolor{darkgreen}{rgb}{0,0.7,0}
\definecolor{darkyellow}{rgb}{0.8,0.8,0}

\definecolor{color_unlabled}{rgb}{0.0,0.0,0.0}
\definecolor{color_vehicle}{rgb}{0.0,0.0,0.56}
\definecolor{color_road}{rgb}{0.5,0.25,0.5}
\definecolor{color_redlight}{rgb}{1.0,0.0,0.0}
\definecolor{color_person}{rgb}{0.859,0.078,0.234}
\definecolor{color_roadline}{rgb}{0.613,0.914,0.195}
\definecolor{color_sidewalk}{rgb}{0.953,0.137,0.906}

\newcommand{\red}[1]{\noindent{{#1}}}

\begin{document}

\title{TransFuser: Imitation with Transformer-Based Sensor Fusion for Autonomous Driving}

\author{Kashyap Chitta,
        Aditya Prakash,
        Bernhard Jaeger,
        Zehao Yu,
        Katrin Renz,
        and Andreas Geiger%
\IEEEcompsocitemizethanks{\IEEEcompsocthanksitem K. Chitta, B. Jaeger, Z. Yu, K. Renz and A. Geiger are with the Autonomous Vision Group, University of T\"{u}bingen and Max Planck Institute for Intelligent Systems, T\"{u}bingen. E-mail: kashyap.chitta@tue.mpg.de, bernhard.jaeger@uni-tuebingen.de,  zehao.yu@uni-tuebingen.de, katrin.renz@uni-tuebingen.de, a.geiger@uni-tuebingen.de.
\IEEEcompsocthanksitem A. Prakash is with the University of Illinois Urbana-Champaign. Work done while at the Max Planck Institute for Intelligent Systems, T\"{u}bingen. E-mail: adityap9@illinois.edu}%
}

\markboth{}%
{Chitta \MakeLowercase{\textit{et al.}}: TransFuser: Imitation with Transformer-Based Sensor Fusion for Autonomous Driving}

\IEEEtitleabstractindextext{%
\begin{abstract}
\justifying How should we integrate representations from complementary sensors for autonomous driving? Geometry-based fusion has shown promise for perception (e.g. object detection, motion forecasting). However, in the context of end-to-end driving, we find that imitation learning based on existing sensor fusion methods underperforms in complex driving scenarios with a high density of dynamic agents. Therefore, we propose TransFuser, a mechanism to integrate image and LiDAR representations using self-attention. Our approach uses transformer modules at multiple resolutions to fuse perspective view and bird's eye view feature maps. We experimentally validate its efficacy on a challenging new benchmark with long routes and dense traffic, as well as the official leaderboard of the CARLA urban driving simulator. At the time of submission, TransFuser outperforms all \red{prior work} on the CARLA leaderboard in terms of \red{driving score} by a large margin. Compared to geometry-based fusion, TransFuser reduces the average collisions per kilometer by \red{48\%}.
\end{abstract}

}
\maketitle
\IEEEdisplaynontitleabstractindextext
\IEEEpeerreviewmaketitle
\IEEEraisesectionheading{\section{Introduction}\label{sec:intro}}

\IEEEPARstart{L}iDAR sensors provide accurate 3D information for autonomous vehicles. While LiDAR-based methods have recently shown impressive results for end-to-end driving~\cite{Rhinehart2020ICLR, Filos2020ICML,Sadat2020ECCV,Cui2021ICCV}, they are evaluated in settings that assume access to privileged information not available through the LiDAR. This includes test-time access to HD maps and ground truth traffic light states. In practice, the information missing in the LiDAR must be recovered from other sensors on the vehicle, such as RGB cameras~\cite{Dosovitskiy2017CORL, Richter2016ECCV, Gaidon2016CVPR, Xu2017CVPRa, Cordts2016CVPR, Geiger2012CVPR, Ros2016CVPR, Waymo2019, Yu2018ARXIV}.

This raises important questions: \textit{Can we integrate representations from these two modalities to exploit their complementary advantages for autonomous driving? To what extent should we process the different modalities independently, and what kind of fusion mechanism should we employ for maximum performance gain?} Prior works in the field of {sensor fusion} have mostly focused on the perception aspect of driving, \eg 2D and 3D object detection~\cite{Fadadu2020ARXIV, Chen2017CVPR, Zhou2019CORL, Chen2020ARXIV, Qi2018CVPR, Ku2018IROSa, Liang2018ECCV, You2020ICLR, Liang2019CVPR, Meyer2019CVPRW}, motion forecasting~\cite{Fadadu2020ARXIV, Luo2018CVPR, Casas2020ICRA, Liang2020CVPR, Zhang2020CVPR, Casas2018CORL, Djuric2020ARXIV, Meyer2020ARXIV, Li2020IROS, Chen2020ARXIV}, and depth estimation~\cite{Fu2020ARXIV, Xu2019ICCV, You2020ICLR, Liang2019CVPR}. These methods focus on learning a state representation that captures the geometric and semantic information of the 3D scene. They operate primarily based on geometric feature projections between the image space and different LiDAR projection spaces, \eg Bird's Eye View (BEV)~\cite{Fadadu2020ARXIV, Chen2017CVPR, Zhou2019CORL, Chen2020ARXIV, Qi2018CVPR, Ku2018IROSa, Liang2018ECCV, You2020ICLR, Liang2019CVPR} and Range View (RV)~\cite{Meyer2019CVPR, Meyer2019CVPRW, Fadadu2020ARXIV, Meyer2020ARXIV, Chen2020ARXIV, Sobh2018NEURIPSW}. Information is typically aggregated from a local neighborhood around each feature in the projected 2D or 3D space.

\begin{figure}
    \centering
    \includegraphics[width=\columnwidth]{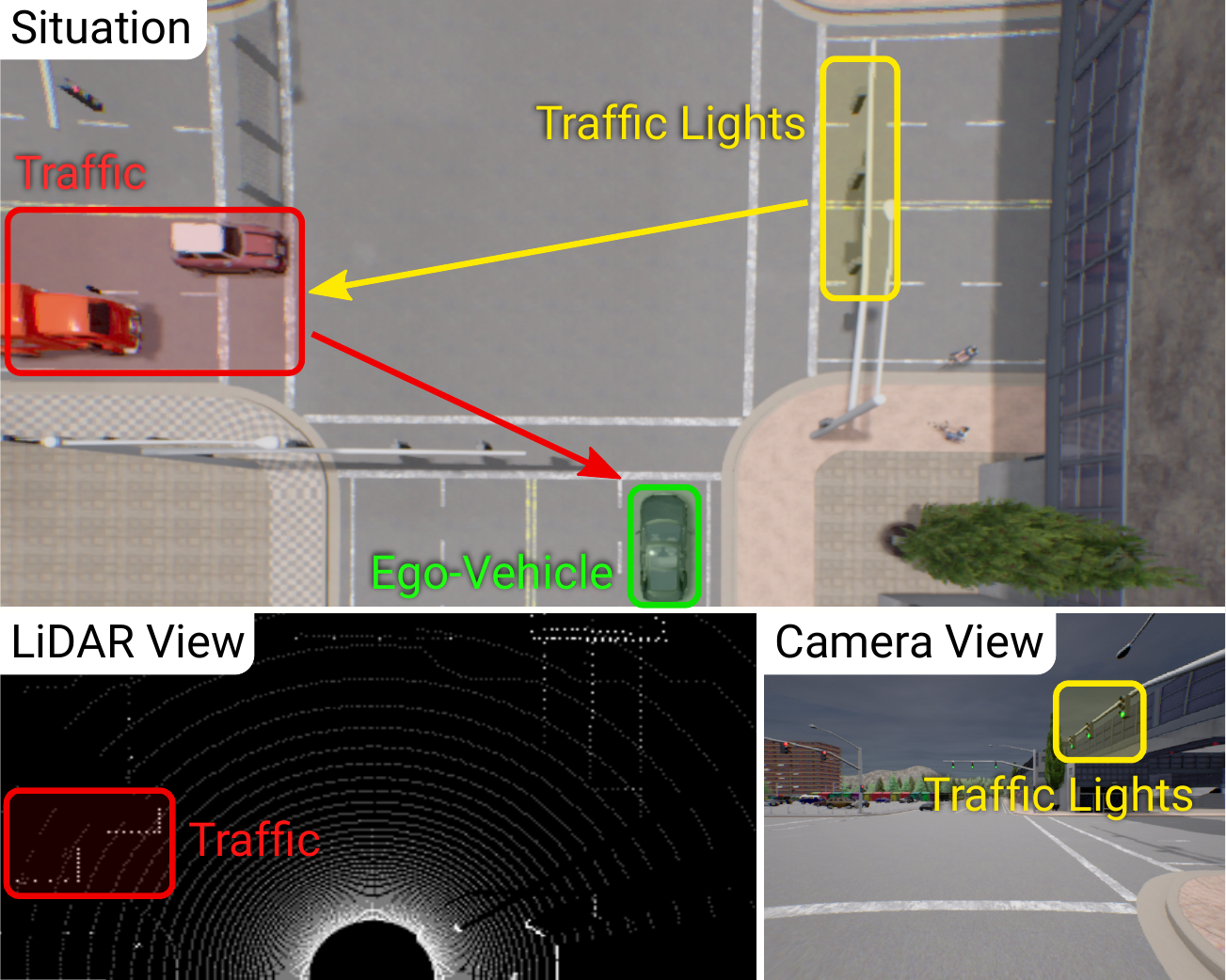}
    \caption{\textbf{Illustration.} Consider an intersection with oncoming traffic from the left. To safely navigate the intersection, the agent ({\color{darkgreen}{green}}) must capture the global context of the scene involving the interaction between the traffic light ({\color{darkyellow}{yellow}}) and the crossing traffic ({\color{red}{red}}). Our TransFuser model integrates geometric and semantic information across multiple modalities via attention mechanisms to capture global context, leading to safe driving behavior in CARLA.}
    \label{fig:teaser}
    \vspace{-0.25cm}
\end{figure}

\begin{figure*}
\centering
\includegraphics[width=\textwidth]{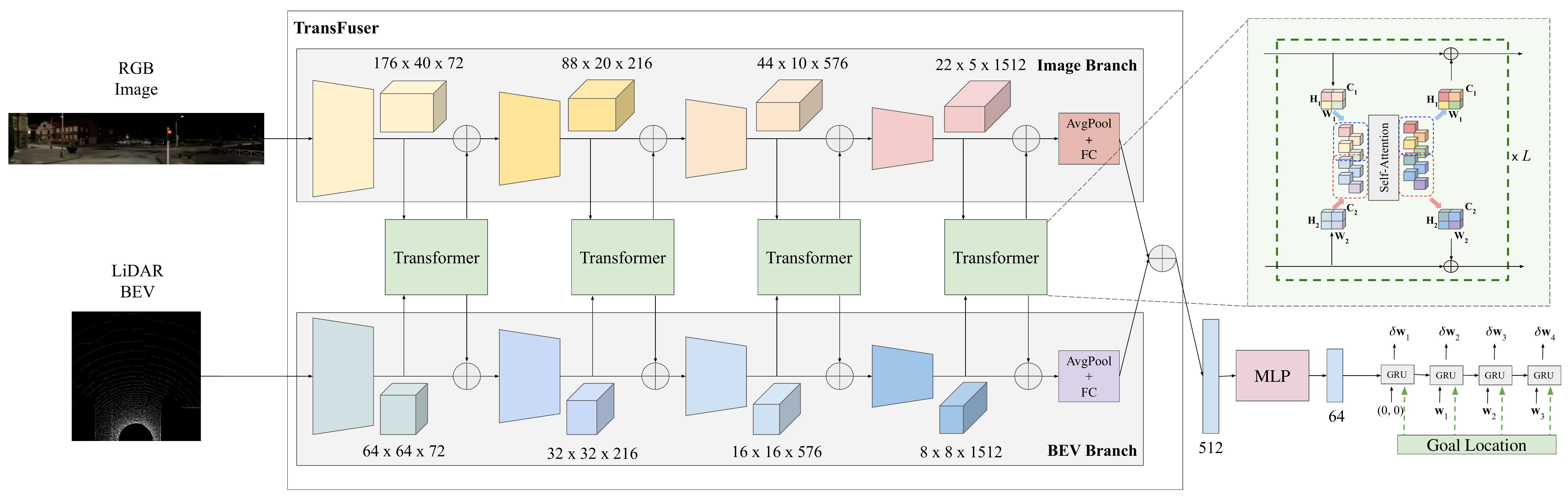}
\caption{\textbf{Architecture.} We consider RGB image and LiDAR BEV representations (\secref{sec:io_parameterization}) as inputs to our multi-modal fusion transformer (TransFuser) which uses several transformer modules for the fusion of intermediate feature maps between both modalities. This fusion is applied at multiple resolutions throughout the feature extractor, resulting in a 512-dimensional feature vector output from both the image and LiDAR BEV stream, which are combined via element-wise summation. This 512-dimensional feature vector constitutes a compact representation of the environment that encodes the global context of the 3D scene. It is then processed with an MLP before passing it to an auto-regressive waypoint prediction network. We use a single layer GRU followed by a linear layer that takes in the hidden state and predicts the differential ego-vehicle waypoints $\{ \delta \bw_{t}\}_{t=1}^T$, represented in the ego-vehicle's current coordinate frame.}
\label{fig:model}
\vspace{-0.0cm}
\end{figure*}

We observe that the locality assumption in these architecture designs hampers performance in complex urban scenarios (\tabref{tab:detailed_results}). For example, when handling traffic at intersections with multiple lanes, the ego-vehicle needs to account for interactions between nearby dynamic agents and traffic lights that are farther away. While deep convolutional networks can be used to capture global context within a single modality, it is non-trivial to extend them to multiple modalities or model interactions between pairs of features. To overcome these limitations, we use the attention mechanism of transformers~\cite{Vaswani2017NEURIPS,Dosovitskiy2021ICLR} to tightly integrate global contextual reasoning about the 3D scene directly into the feature extraction layers of different modalities. We consider image and LiDAR inputs since they are complementary to each other, and focus on integrating representations between these modalities (\figref{fig:teaser}). The inputs are processed by two independent convolutional encoder branches, which are interconnected using transformers (\figref{fig:model}). We call the resulting model \emph{TransFuser} and integrate it into an auto-regressive waypoint prediction framework designed for end-to-end driving.

To show the advantages of our approach, we conduct a comprehensive study using the CARLA driving simulator~\cite{Dosovitskiy2017CORL}. We consider a more challenging evaluation setting than existing closed-loop driving benchmarks (\eg NoCrash benchmark~\cite{Codevilla2019ICCV}, NEAT routes~\cite{Chitta2021ICCV}) based on the new CARLA version 0.9.10 leaderboard~\cite{Leaderboard2020}. Our proposed \textit{Longest6} benchmark involves $\sim$1.5km long routes, increased traffic density, and challenging pre-crash traffic scenarios. To tackle these challenges, we incorporate auxiliary supervision signals in a multi-task learning setup to train TransFuser and several strong baselines. On both the proposed benchmark and the secret routes of the official CARLA leaderboard, TransFuser \red{achieves a significantly higher driving score than prior work}.

Our contributions can be summarized as follows:
\begin{itemize}
    \item We design a new evaluation setting in CARLA which demonstrates that imitation learning policies based on existing sensor fusion approaches are unable to handle challenging scenarios with dense traffic.
    \item We propose a novel multi-modal fusion transformer (TransFuser) to incorporate global context and pair-wise interactions into the feature extraction layers of different input modalities.
    \item We conduct a detailed empirical analysis demonstrating state-of-the-art driving performance with TransFuser on both the proposed evaluation setting and the official CARLA leaderboard. Our analysis provides insights and explores the current limitations of end-to-end driving models.
\end{itemize}

This journal paper is an extension of a conference paper published at CVPR 2021~\cite{Prakash2021CVPR}: we enhance the TransFuser model from \cite{Prakash2021CVPR} to obtain state-of-the-art performance by incorporating (1) an improved expert demonstrator for data collection, (2) a new sensor configuration with an increased field of view through multiple cameras, (3) an improved vision backbone architecture, and (4) an updated training procedure involving multi-task learning. \red{We also provide a new image-only baseline, \textit{Latent TransFuser}, which significantly outperforms prevalent baselines used for CARLA. Our updated code, dataset, and trained models are available at} \url{https://github.com/autonomousvision/transfuser}.
\section{Related Work}
\label{sec:related}

\boldparagraph{Multi-Modal Autonomous Driving} Recent multi-modal methods for end-to-end driving~\cite{Xiao2019ARXIV, Behl2020IROS, Zhou2019SR, Sobh2018NEURIPSW, Natan2022ARXIV} have shown that complementing RGB images with depth and semantics has the potential to improve driving performance. Xiao et al.~\cite{Xiao2019ARXIV} explore RGBD input from the perspective of early, mid and late fusion of camera and depth modalities and observe significant gains. Behl et al.~\cite{Behl2020IROS} and Zhou et al.~\cite{Zhou2019SR} demonstrate the effectiveness of semantics and depth as explicit intermediate representations for driving. \red{Natan et al.~\cite{Natan2022ARXIV} combine 2D semantics and depth into a semantic point cloud for end-to-end driving.} In this work, we focus on image and LiDAR inputs since they are complementary to each other in terms of representing the scene and are readily available in autonomous driving systems. In this respect, Sobh et al.~\cite{Sobh2018NEURIPSW} exploit a late fusion architecture for LiDAR and image modalities where each input is encoded in a separate stream from which the final feature vectors are concatenated together. \red{The concurrent work of Chen et al.~\cite{Chen2022CVPR} performs sensor fusion via PointPainting, which concatenates semantic class information extracted from the RGB image to the LiDAR point cloud~\cite{Vora2020CVPR}. We observe that the late fusion mechanism of Sobh et al.~\cite{Sobh2018NEURIPSW} suffers from high infraction rates and the PointPainting-based system has a reduced route completion percentage in comparison to several baselines (\tabref{tab:detailed_results}).} To mitigate these limitations, we propose a multi-modal fusion transformer (TransFuser) that is effective in integrating information from different modalities at multiple stages during feature encoding using attention. TransFuser helps to capture the global context of the 3D scene which requires fusing information from distinct spatial locations across sensors.

\boldparagraph{Sensor Fusion Methods for Object Detection and Motion Forecasting} Most sensor fusion works consider perception tasks, \eg object detection~\cite{Fadadu2020ARXIV, Chen2017CVPR, Zhou2019CORL, Chen2020ARXIVa, Qi2018CVPR, Ku2018IROSa, Liang2018ECCV, You2020ICLR, Liang2019CVPR, Vora2020CVPR, Meyer2019CVPRW, Laddha2021CVPRW, Gautam2021ICCVW, Dai2021, Mohta2021ARXIV, Mees2016IROS, Xu2018CVPR, Qi2018CVPRb, Pang2020IROS, Zhao2019AAAI, Bai2022CVPR, Chen2022ARXIV, Liu2022ARXIVb} and motion forecasting~\cite{Luo2018CVPR, Casas2020ICRA, Liang2020CVPR, Zhang2020CVPR, Casas2018CORL, Djuric2020ARXIV, Meyer2020ARXIV, Laddha2021CVPRW, Khalil2021IEEEA, Wang2022ARXIV}. They operate on multi-view LiDAR, \eg Bird's Eye View (BEV) and Range View (RV), or complement the camera input with depth information from LiDAR. This is typically achieved by projecting LiDAR features into the image space or projecting image features into the BEV or RV space. The closest approach to ours is ContFuse~\cite{Liang2018ECCV} which performs multi-scale dense feature fusion between image and LiDAR BEV features. For each pixel in the LiDAR BEV representation, it computes the nearest neighbors in a local neighborhood in 3D space, projects these neighboring points into the image space to obtain the corresponding image features, aggregates these features using continuous convolutions, and combines them with the LiDAR BEV features. \red{Concurrent to our work, EPNET++~\cite{Liu2021ARXIVb} and CAT-Det~\cite{Zhang2022CVPR} also employ multi-scale bidirectional fusion between image and LiDAR point clouds using attention to learn enhanced feature representations for 3D object detection.} Other projection-based fusion methods follow a similar trend and aggregate information from a local neighborhood in 2D or 3D space. However, the state representation learned by these methods is insufficient since they do not capture the global context of the 3D scene, which is important for safe maneuvers in dense traffic. To demonstrate this, we implement a multi-scale geometry-based fusion mechanism, inspired by~\cite{Liang2018ECCV, Liang2019CVPR}, involving both image-to-LiDAR and LiDAR-to-image feature fusion for end-to-end driving and observe high infraction rates in the dense urban setting (\tabref{tab:detailed_results}).

\boldparagraph{Attention for Autonomous Driving} Attention has been explored in the context of driving for lane changing~\cite{Chen2019IROS}, object detection~\cite{Chen2017ARXIVa, Li2020IROS, Yuan2021IEEE, Sheng2021CVPR}, motion forecasting~\cite{Li2020IROS, Sadeghian2018ECCV, Sadeghian2019CVPR, Huang2019ICCV, Choi2019ICCV, Kosaraju2019NEURIPS, Ivanovic2019ICCV, Wei2021ICRA, Yuan2021ARXIV, Ngiam2021ARXIV}, driver attention prediction~\cite{Gou2022SC, Cao2022ARXIV} and recently also for end-to-end driving~\cite{Prakash2021CVPR, Chitta2021ICCV}. Chen et al.~\cite{Chen2017ARXIVa} employ a recurrent attention mechanism over a learned semantic map for predicting vehicle controls. Li et al.~\cite{Li2020IROS} utilize attention to capture temporal and spatial dependencies between actors by incorporating a transformer module into a recurrent neural network. SA-NMP~\cite{Wei2021ICRA} learns an attention mask over features extracted from a 2D CNN, operating on LiDAR BEV projections and HD maps, to focus on dynamic agents for safe motion planning. Chen et al.~\cite{Chen2019IROS} utilize spatial and temporal attention in a hierarchical deep reinforcement learning framework to focus on the surrounding vehicles for lane changing in the TORCS simulator. PYVA~\cite{Yang2021CVPR} uses attention to perform image translation from the perspective view to the BEV. \red{This idea has also been adopted by several concurrent papers on BEV semantic segmentation from image inputs~\cite{Saha2022ICRA, Zhou2022CVPR, Peng2022ARXIV, Xie2022ARXIV, Li2022ARXIV}.} NEAT~\cite{Chitta2021ICCV} uses intermediate attention maps to iteratively compress high dimensional 2D image features into a compact BEV representation for driving. Compared to NEAT, our attention mechanism is simpler since it does not require iterative refinement of the attention at test-time. Unlike NEAT, we also apply our attention mechanism at multiple feature resolutions, enabling sensor fusion for both shallow and deep features in the network. Furthermore, none of the existing attention-based driving approaches consider multiple sensor modalities. Our work uses the self-attention mechanism of transformers for dense fusion of image and LiDAR features.

\section{TransFuser}
\label{sec:method}

In this work, we propose a novel architecture for end-to-end driving (\figref{fig:model}). It has two main components: (1) a multi-modal fusion transformer for integrating information from multiple sensor modalities (image and LiDAR), and (2) an auto-regressive waypoint prediction network. The following sections detail our problem setting, input and output parameterization, and each component of the model.

\subsection{Problem Setting}
\label{sec:problem_setting}
We consider the task of point-to-point navigation in an urban setting~\cite{Filos2020ICML, Rhinehart2019ICCV, Rhinehart2020ICLR, Chen2019CORL, Codevilla2019ICCV} where the goal is to complete a given route while safely reacting to other dynamic agents and following traffic rules.

\boldparagraph{Imitation Learning (IL)} The goal of IL is to learn a policy $\pi$ that imitates the behavior of an expert $\pi^{*}$. In our setup, a policy is a mapping from inputs to waypoints that are provided to a separate low-level controller to output actions. We consider the Behavior Cloning (BC) approach of IL which is a supervised learning method. An expert policy is first rolled out in the environment to collect a dataset, $\cD = \{ (\cX^i , \cW^i) \}_{i=1}^Z$ of size $Z$, which consists of high-dimensional observations of the environment, $\cX$, and the corresponding expert trajectory, defined by a set of 2D waypoints in BEV space, \ie, $\cW = \{ \bw_t = (x_t, y_t) \}_{t=1}^T$. This BEV space uses the coordinate frame of the ego-vehicle. The policy, $\pi$, is trained in a supervised manner using the collected data, $\cD$, with the loss function, $\cL$.
\begin{equation}
    \argmin_{\pi} \nE_{(\cX, \cW) \sim \cD} \left[ \cL (\cW, \pi(\cX)) \right]
\end{equation}
The high-dimensional observation, $\cX$, includes a front camera image input and a LiDAR point cloud from a single time-step. We use a single time-step input since prior works on IL for autonomous driving have shown that using observation histories may not lead to performance gain~\cite{Muller2005NEURIPS, Wang2019IROS, Bansal2019RSS, Wen2020NEURIPS, Wen2021ICML}. \red{For $\cL$, we use} the $L_1$ distance between the predicted trajectory, $\pi(\cX)$, and the expert trajectory, $\cW$, as the primary loss function. Furthermore, we use several auxiliary losses used to boost performance, which are detailed in \secref{sec:loss}. We assume access to an inverse dynamics model~\cite{Bellman2015Princeton}, implemented as a PID Controller $\nI$, which performs the low-level control, \ie, steer, throttle, and brake, provided the future trajectory $\cW$. The actions are determined as $\ba = \nI (\cW)$.

\boldparagraph{Global Planner} We follow the standard protocol of CARLA 0.9.10 and assume that high-level goal locations $\cG$ are provided as GPS coordinates. Note that these goal locations are sparse and can be hundreds of meters apart, as opposed to the local waypoints predicted by the policy $\pi$.

\subsection{Input and Output Parameterization}
\label{sec:io_parameterization}

\boldparagraph{Input Representation} Following previous LiDAR-based driving approaches~\cite{Rhinehart2019ICCV, Filos2020ICML}, we convert the LiDAR point cloud into a 2-bin histogram over a 2D BEV grid with a fixed resolution. We consider the points within 32m in front of the ego-vehicle and 16m to each of the sides, thereby encompassing a BEV grid of 32m $\times$ 32m. We divide the grid into blocks of 0.125m $\times$ 0.125m which results in a resolution of 256 $\times$ 256 pixels. For the histogram, we discretize the height dimension into 2 bins representing the points on/below and above the ground plane. We also rasterize the 2D goal location in the same 256 $\times$ 256 BEV space as the LiDAR point cloud and concatenate this channel to the 2 histogram bins. This results in a three-channel pseudo-image of size $256 \times 256$ pixels. We represent the goal location in the BEV as this correlates better with the waypoint predictions compared to the perspective image domain~\cite{Chen2019CORL}.

For the RGB input, we use three cameras (facing forward, 60$^{\circ}$ left and 60$^{\circ}$ right). Each camera has a horizontal FOV of 120$^{\circ}$. We extract the images at a resolution of 960 $\times$ 480 pixels, which we crop to 320 $\times$ 160 to remove radial distortion at the edges. These three undistorted images are composed into a single image input to the encoder, which has a resolution of 704 $\times$ 160 pixels and 132$^{\circ}$ FOV. We find that this FOV is sufficient to observe both near and far traffic lights in all public towns of CARLA.

\boldparagraph{Output Representation} We predict the future trajectory $\cW$ of the ego-vehicle in BEV space, centered at the current coordinate frame of the ego-vehicle. The trajectory is represented by a sequence of 2D waypoints, $\{\bw_t = (x_t, y_t)\}_{t=1}^T$. We use $T=4$, which is the default number of waypoints required by our inverse dynamics model.

\subsection{Multi-Modal Fusion Transformer}
\label{sec:arch}

Our key idea is to exploit the self-attention mechanism of transformers~\cite{Vaswani2017NEURIPS} to incorporate the global context for image and LiDAR modalities, given their complementary nature. The transformer architecture takes as input a sequence consisting of discrete tokens, each represented by a feature vector. The feature vector is supplemented by a positional encoding to incorporate spatial inductive biases. 

Formally, we denote the input sequence as $\bF^{in} \in \nR^{N \times D_f}$, where $N$ is the number of tokens in the sequence, and each token is represented by a feature vector of dimensionality $D_f$. The transformer uses linear projections for computing a set of queries, keys, and values ($\bQ$, $\bK$, and $\bV$),
\begin{equation}
    \bQ = \bF^{in}\bM^q, \, \, \bK = \bF^{in}\bM^k, \, \, \bV = \bF^{in}\bM^v
\label{eqn:transformer}
\end{equation}
where $\bM^q \in \nR^{D_f \times D_q}$, $\bM^k \in \nR^{D_f \times D_k}$ and $\bM^v \in \nR^{D_f \times D_v}$ are weight matrices. It uses the scaled dot products between $\bQ$ and $\bK$ to compute the attention weights and then aggregates the values for each query,
\begin{equation}
    \bA = \text{softmax} \bigg(\frac{\bQ \bK^T} {\sqrt{D_k}}\bigg) \bV
\label{eqn:transformer_attention}
\end{equation}
Finally, the transformer uses a non-linear transformation to calculate the output features, $\bF^{out}$ which are of the same shape as the input features, $\bF^{in}$.
\begin{equation} \label{eqn:transformer_mlp}
    \bF^{out} = \text{MLP}(\bA) + \bF^{in}
\end{equation}
The transformer applies the attention mechanism multiple times throughout the architecture, resulting in $L$ attention layers. Each layer in a standard transformer has multiple parallel attention `heads', which involve generating several $\bQ$, $\bK$ and $\bV$ values per $\bF^{in}$ for Eq. \eqref{eqn:transformer} and concatenating the resulting values of $\bA$ from Eq. \eqref{eqn:transformer_attention}.

Unlike the token input structures in NLP, we operate on grid structured feature maps. Similar to prior works on the application of transformers to images~\cite{Sun2019ICCV,Chen2020ICML,Qi2020ARXIV,Dosovitskiy2020ARXIV}, we consider the intermediate feature maps of each modality to be a set rather than a spatial grid and treat each element of the set as a token. The convolutional feature extractors for the image and LiDAR BEV inputs encode different aspects of the scene at different layers. Therefore, we fuse these features at multiple scales (\figref{fig:model}) throughout the encoder.

Let the intermediate grid structured feature map of a single modality indexed by $s$ be a 3D tensor of dimension $H_s \times W_s \times C$. For $S$ different modalities, these features are stacked together to form a sequence of dimension $\sum_{s=1}^S(H_s*W_s) \times C$. We add a learnable positional embedding, which is a trainable parameter of the same dimension as the stacked sequence, so that the network can infer spatial dependencies between different tokens at train time. The input sequence and positional embedding are combined using element-wise summation to form a tensor of dimension $\sum_{s=1}^S(H_s*W_s) \times C$. As shown in \figref{fig:model}, this tensor is fed as input to the transformer, which produces an output of the same dimension. We have omitted the positional embedding and velocity embedding inputs in \figref{fig:model} for clarity. The output is then reshaped into $S$ feature maps of dimension $H_s \times W_s \times C$ each and fed back into each of the individual modality branches using an element-wise summation with the existing feature maps. The mechanism described above constitutes feature fusion at a single scale. This fusion is applied \textit{multiple times} throughout the feature extractors of the image and BEV branches at different resolutions (\figref{fig:model}). However, processing feature maps at high spatial resolutions is computationally expensive. Therefore, we downsample higher resolution feature maps from the early encoder blocks using average pooling to the same resolution as the final feature map before passing them as inputs to the transformer. We upsample the output to the original resolution using bilinear interpolation before element-wise summation with the existing feature maps.

After carrying out dense feature fusion at multiple resolutions (\figref{fig:model}), we obtain a feature map of dimensions $22 \times 5 \times C$ from the feature extractors of the image branch, and $8 \times 8 \times C$ from the BEV branch. Where $C$ is the number of channels at the current resolution in the feature extractor and lies in $\{72,216,576,1512\}$. These feature maps are reduced to a dimension of 512 by average pooling, followed by a fully-connected layer of 512 units. The feature vector of dimension 512 from both the image and the LiDAR BEV streams are then combined via element-wise summation. This 512-dimensional feature vector constitutes a compact representation of the environment that encodes the global context of the 3D scene. This is then fed to the waypoint prediction network, which we describe next.

\subsection{Waypoint Prediction Network}
\label{sec:decoder}

As shown in~\figref{fig:model}, we pass the 512-dimensional feature vector through an MLP (comprising 2 hidden layers with 256 and 128 units) to reduce its dimensionality to 64 for computational efficiency before passing it to the auto-regressive waypoint network implemented using GRUs~\cite{Cho2014EMNLP}. We initialize the hidden state of the GRU with the 64-dimensional feature vector. The update gate of the GRU controls the flow of information encoded in the hidden state to the output and the next time-step. It also takes in the current position and the goal location (\secref{sec:problem_setting}) as input, which allows the network to focus on the relevant context in the hidden state for predicting the next waypoint. We provide the GPS coordinates of the goal location (registered to the ego-vehicle coordinate frame) as input to the GRU in addition to the encoder since it can more directly influence the waypoint predictions. Following~\cite{Filos2020ICML}, we use a single layer GRU followed by a linear layer which takes in the hidden state and predicts the differential ego-vehicle waypoints $\{ \delta \bw_{t}\}_{t=1}^T$ for $T=4$ future time-steps in the ego-vehicle current coordinate frame. Therefore, the predicted future waypoints are given by $\{ \bw_t = \bw_{t-1} + \delta \bw_t \}_{t=1}^T$. The input to the first GRU unit is given as (0,0) since the BEV space is centered at the ego-vehicle's position.

\subsection{Controller}
\label{sec:control}

We use two PID controllers for lateral and longitudinal control to obtain steer, throttle, and brake values from the predicted waypoints, $\{ \bw_t \}_{t=1}^T$. The longitudinal controller takes in the magnitude of a weighted average of the vectors between waypoints of consecutive time steps, whereas the lateral controller takes in their orientation. For the PID controllers, we use the same configuration as in the author-provided codebase of~\cite{Chen2019CORL}. Additional details regarding the controllers can be found in the supplementary material.

\boldparagraph{Creeping} If the car has not moved for a long duration (55 seconds, chosen to be higher than the expected wait time at a red light), we make it creep forward by setting the target speed of the PID controller to 4 m/s for a short time (1.5 seconds). This creeping behavior is used to recover from the \textit{inertia problem} observed in IL for autonomous driving~\cite{Codevilla2019ICCV}. When a vehicle is still, the probability that it continues to stay in place (\eg in dense traffic) is very high in the training data. This can lead to the trained agent never starting to drive again after having stopped.

\boldparagraph{Safety Heuristic} The creeping behavior alone would be unsafe, \eg in situations where the agent is stuck in traffic where creeping forward could lead to a collision. To prevent this, we implement a safety check that overwrites the creeping behavior if there are any LiDAR hits in a small rectangular area in front of the car. \red{While this heuristic is essential during creeping, it can also be applied during regular driving to enhance the safety~\cite{Vitelli2021ARXIV}. We study the impact of applying a global safety heuristic during both creeping and regular driving in \secref{sec:safety}.}

\subsection{Loss Functions}
\label{sec:loss}

Following~\cite{Chen2019CORL}, we train the network using an $L_1$ loss between the predicted waypoints and the ground truth waypoints (from the expert), registered to the current coordinate frame. Let $\bw_t^{gt}$ represent the ground truth waypoint for time-step $t$, then the loss function is given by:
\begin{equation}
\label{eqn:loss}
    \cL = \sum_{t=1}^{T} {||\bw_t - \bw_t^{gt}||}_1
\end{equation}
Note that the ground truth waypoints $\{\bw_t^{gt}\}$ which are available only at training time are different from the sparse goal locations $\cG$ provided at both training and test time.

\begin{figure}
    \centering
    \includegraphics[width=\columnwidth]{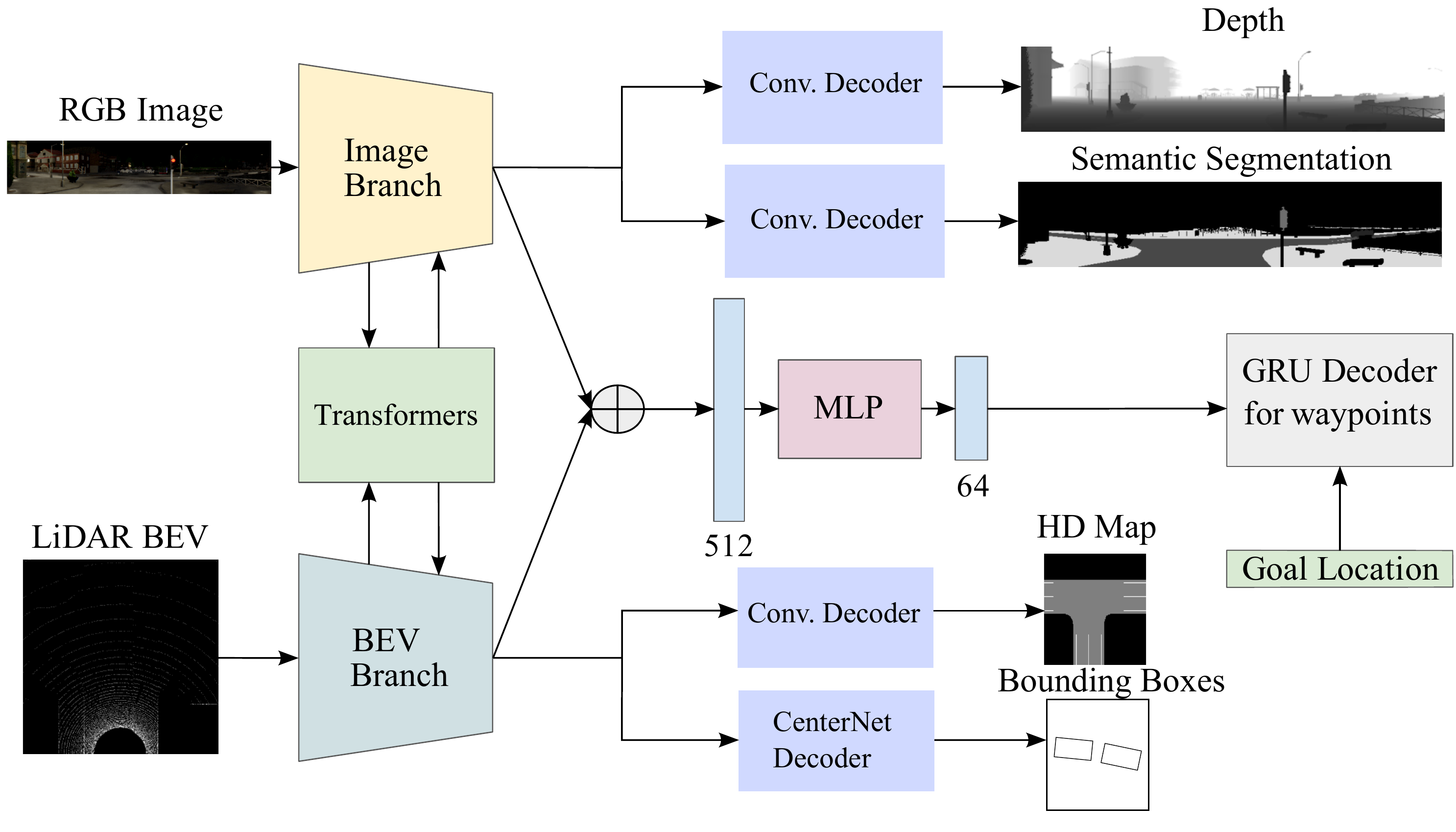}
    \caption{\textbf{Auxiliary Loss Functions.} Besides the waypoint loss (Eq. \eqref{eqn:loss}), we incorporate four auxiliary tasks: depth prediction and semantic segmentation from the image branch; HD map prediction and vehicle detection from the BEV branch.}
    \label{fig:aux}
    \vspace{-0.0cm}
\end{figure}

\boldparagraph{Auxiliary Tasks} To account for the complex spatial and temporal scene structure encountered in autonomous driving, the training objectives used in IL-based driving agents have evolved by incorporating auxiliary tasks. Training signals aiming to reconstruct the scene have become common, such as image auto-encoding~\cite{Ohn-Bar2020CVPR}, 2D semantic segmentation~\cite{Huang2020SENSORS}, Bird's Eye View (BEV) semantic segmentation~\cite{Loukkal2020ARXIV}, 2D semantic prediction~\cite{Hu2020ECCV}, and BEV semantic prediction~\cite{Sadat2020ECCV,Chitta2021ICCV}. Performing auxiliary tasks has been shown to lead to more interpretable and robust models~\cite{Sadat2020ECCV,Chitta2021ICCV}. In this work, we consider four auxiliary tasks: depth prediction, semantic segmentation, HD map prediction, and vehicle detection (\figref{fig:aux}).

\boldparagraph{2D Depth and Semantics} Combining 2D depth estimation and 2D semantic segmentation as auxiliary tasks has been an effective approach for image-based end-to-end driving~\cite{Li2018ARXIV,Chitta2021ICCV}. We use the same decoder architecture as the AIM-MT baseline of \cite{Chitta2021ICCV} to decode depth and semantics from the image branch features. The depth output is supervised with an $L_1$ loss, and the semantics with a cross-entropy loss. Following \cite{Chitta2021ICCV}, we consider 7 semantic classes: (1) unlabeled, (2) vehicle, (3) road, (4) red light, (5) pedestrian, (6) lane marking, and (7) sidewalk.

\boldparagraph{HD Map} We predict a three-channel BEV segmentation mask containing the classes road, lane marking and other. This encourages the intermediate features to encode information regarding drivable and non-drivable areas. The map uses the same coordinate frame as the LiDAR input, and is therefore obtained from the feature map of the LiDAR branch with a convolutional decoder. However, we predict a downsampled version of the HD map ($64 \times 64$ instead of $256 \times 256$) for computational efficiency. The HD map prediction task uses a cross-entropy loss.

\boldparagraph{Bounding Boxes} We locate other vehicles in the scene via keypoint estimation with a CenterNet decoder~\cite{Zhou2019ARXIVb}. Specifically, we predict a position map $\hat{\bP} \in [0,1]^{64 \times 64}$ from the BEV features using a convolutional decoder. Similar to the HD map prediction task, the $256 \times 256$ input is mapped to smaller $64 \times 64$ predictions for vehicle detection. The 2D target label for this task is rendered with a Gaussian kernel at each object center of our training dataset. \red{While the orientation is a single scalar value, directly regressing this value is challenging, as observed in existing 3D detectors~\cite{Mousavian2017CVPR,Zhou2019ARXIVb,Yang2020CVPRb}. Therefore, our CenterNet implementation takes a two-stage approach of predicting an initial coarse orientation followed by a fine offset angle.} To predict the coarse orientation, we discretize the relative yaw of each ground truth vehicle into 12 bins of size 30$^{\circ}$, and predict this class via a 12-channel classification label at each pixel, $\hat{\bO} \in [0,1]^{64 \times 64 \times 12}$, as in~\cite{Yang2020CVPRb}. Finally, we predict a regression map $\hat{\bR} \in \nR^{64 \times 64 \times 5}$. This regression map holds three regression targets: vehicle size ($\in \nR^2$), position offset ($\in \nR^2$) and orientation offset ($\in \nR$). The position offset is used to make up for quantization errors introduced by predicting position maps at a lower resolution than the inputs. The orientation offset corrects the orientation discretization error~\cite{Yang2020CVPRb}. Note that only locations with vehicle centers are supervised for predicting $\hat{\bO}$ and $\hat{\bR}$. The position map, orientation map and regression map use a focal loss~\cite{Yin2017ICCV}, cross-entropy loss, and $L_1$ loss respectively.

\subsection{Latent TransFuser}
\label{sec:ltf}

{CILRS}~\cite{Codevilla2019ICCV} is a widely used image-only baseline for the old CARLA version 0.8.4. It learns to directly predict vehicle controls (as opposed to waypoints) from visual features while being conditioned on a discrete navigational command (follow lane, change lane left/right, turn left/right). However, as shown in~\cite{Chitta2021ICCV} this approach obtains poor results on the challenging CARLA leaderboard. Despite this, recent studies involving image-based driving on CARLA have shifted from IL towards more complex Reinforcement Learning (RL) based training, while using CILRS as the primary IL baseline~\cite{Toromanoff2020CVPR,Chen2021ICCVb}. To provide a more meaningful IL baseline for future studies, we now introduce an image-only version of our approach, called Latent TransFuser.

Latent TransFuser replaces the 2-channel LiDAR BEV histogram input to our architecture with a 2-channel positional encoding of identical dimensions ($256 \times 256 \times 2$). The 2D positional encoding is a grid with equally-spaced values from -1 to 1, with one channel for the left-right axis, and one for the top-down axis. Other than this change, the architecture, training procedure, and auxiliary losses remain identical to the LiDAR-based TransFuser. \red{Additionally, our controller uses the LiDAR input for its safety heuristic while creeping (\secref{sec:control}). For Latent TransFuser, we check for an intersection between the small rectangular safety area in front of the car and any detected object from the auxiliary CenterNet detection head instead. The creeping is disabled whenever such an intersection occurs.}

The positional encoding input of Latent TransFuser acts as a proxy for the BEV LiDAR. Since the LiDAR branch is supervised to predict the HD map and bounding boxes, which are in the BEV coordinate frame, fusing image features with the latent features in this branch acts as an attention-based projection from the perspective view to the BEV. The architecture shares similarities with existing attention-based camera to BEV projection techniques such as NEAT~\cite{Chitta2021ICCV} and PYVA~\cite{Yang2021CVPR}. However, unlike these methods, Latent TransFuser projects features at multiple feature resolutions. We show in our experiments that Latent TransFuser is a powerful baseline, outperforming far more complex RL-based methods on the CARLA leaderboard in the image-only input setting (\tabref{tab:leaderboard}).
\section{Experiments}
\label{sec:results}

In this section, we describe our experimental setup, compare the \textbf{driving performance} of TransFuser against several baselines, visualize the \textbf{attention maps} of TransFuser and present \textbf{ablation studies} to highlight the importance of different components of our approach.

\subsection{Task}
\label{sec:task}
We consider the task of navigation along a set of predefined routes in a variety of areas, \eg freeways, urban areas, and residential districts. The routes are defined by a sequence of sparse goal locations in GPS coordinates provided by a global planner. Each route consists of several scenarios, initialized at predefined positions, which test the ability of the agent to handle different kinds of adversarial situations, \eg obstacle avoidance, unprotected turns at intersections, vehicles running red lights, and pedestrians emerging from occluded regions to cross the road at random locations. The agent needs to complete the route within a specified time limit while following traffic regulations and coping with high densities of dynamic agents.

\subsection{Training Dataset}
\label{sec:data}
We use the CARLA~\cite{Dosovitskiy2017CORL} simulator for training and testing, specifically CARLA 0.9.10 which consists of 8 publicly available towns. We use all 8 towns for training. Our dataset is collected along a set of training routes: around 2500 routes through junctions with an average length of 100m, and around 1000 routes along curved highways with an average length of 400m. For generating training data, we roll out an expert policy designed to drive using privileged information from the simulation and store data at 2FPS. The expert is a rule-based algorithm similar to the CARLA traffic manager autopilot\footnote{\url{https://carla.readthedocs.io/en/latest/adv_traffic_manager/}}. Our training dataset contains 228k frames in total. In the following, we provide more details regarding the expert algorithm.

\begin{figure}
    \begin{subfigure}[b]{0.48\textwidth}
        \centering
        \includegraphics[width=\textwidth]{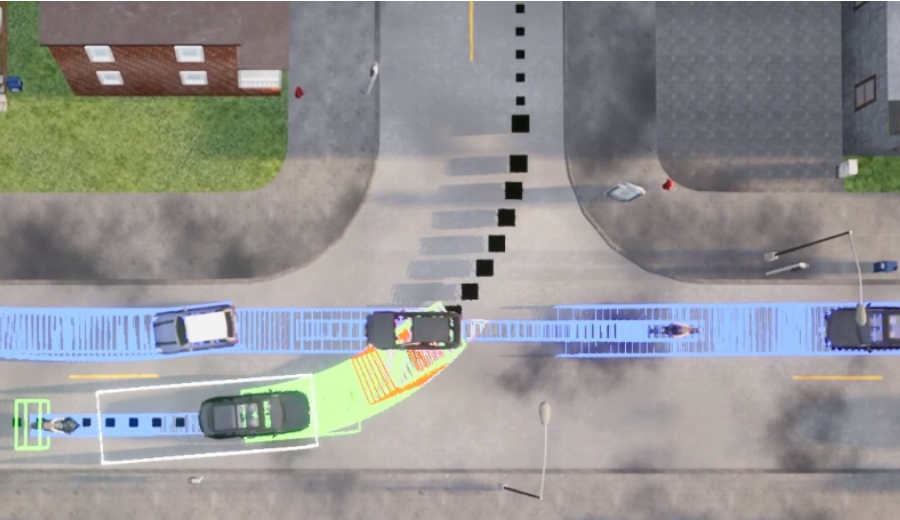}
        \caption{The expert waits before taking the turn because the trajectory forecasting predicts a collision if the expert would drive.}
        \label{fig:expert1}
    \end{subfigure}
    \hfill
    \vspace{0.15cm}
    \begin{subfigure}[b]{0.48\textwidth}
        \centering
        \includegraphics[width=\textwidth]{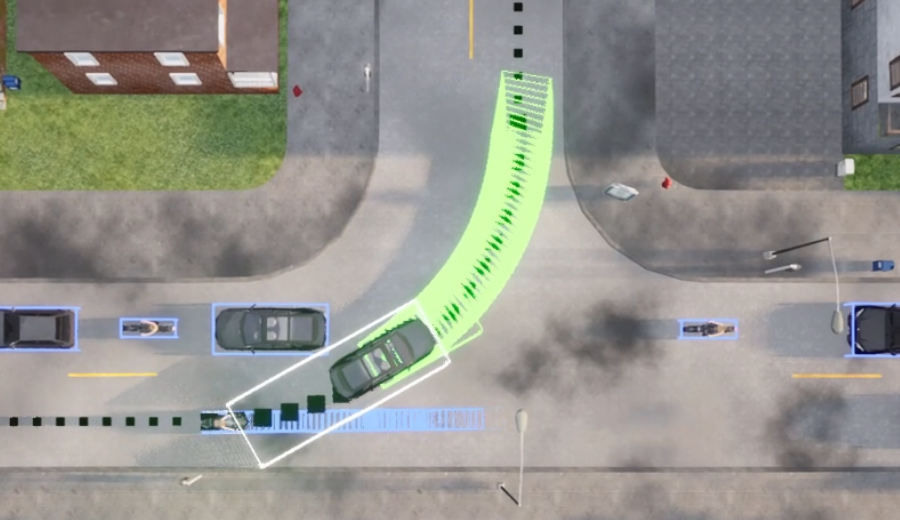}
        \caption{After the oncoming cars have passed, the expert crosses the intersection.}
        \label{fig:expert2}
    \end{subfigure}
    \vspace{0.0cm}
    \caption{\label{fig:expert}\red{\textbf{Expert performing an unprotected left turn.}} The black boxes on the street mark the path that the expert has to follow. The predictions of the bicycle model are colored green for the expert and blue for all other vehicles. Red bounding boxes mark predicted collisions. The white box around the car is used to detect the traffic light trigger boxes that are placed on the street (\eg bottom left \figref{fig:expert1}).}
\end{figure}

\subsection{Expert}
For generating training data, we roll out an expert policy designed to drive using privileged information from the simulator. \red{The waypoints of the expert are the ground truth labels for the imitation loss, so it can be viewed as an automatic labeling algorithm. The ground truth labels for the auxiliary tasks are provided by the simulator.} We build upon the code provided by the authors of~\cite{Chen2019CORL}. This approach is based on simple handcrafted rules. Building the expert with RL is also possible~\cite{Zhang2021ICCV, Agarwal2021ARXIV} but it is more computationally demanding and less interpretable. Our expert policy consists of an A* planner followed by 2 PID controllers (for lateral and longitudinal control). The lateral and longitudinal control tasks are treated independently.

Lateral control is done by following the path generated by the A* planner. Specifically, we minimize the angle of the car towards the next waypoint in the route, which is at least 3.5 meters away, using a PID controller. Longitudinal control is done using a version of model predictive control and differentiates between 3 target speeds. The standard target speed is 4.0 m/s. When the expert is inside an intersection, the target speed is reduced to 3.0 m/s. Lastly, in case an infraction is predicted the target speed is set to 0.0 m/s bringing the vehicle to a halt. We predict red light infractions by performing intersection tests with trigger boxes that CARLA provides. Collision infractions are predicted by forecasting the oriented bounding boxes of all traffic participants. We forecast in 50 ms discrete steps, for 4 seconds in intersections and 1 second in other areas. The forecasting is done using the pretrained \red{kinematic} bicycle model from \cite{Chen2021ICCVb}. \red{This bicycle model is a simple mathematical model that can predict the position, orientation, and speed of a car after a discrete time step, given its current position orientation speed and the applied control. We forecast all vehicles by iteratively rolling out the bicycle model using its output at time step $t$ as the input for time step $t+1$. Since we only know the control input of other traffic participants at the current time step (provided by the simulator), we assume that they will continue to apply the same control at future time steps. For the ego vehicle, we calculate its future steering and throttle by using PID controllers that try to follow the route. The ego brake control is always set to 0 because we want to answer the counterfactual question of whether there will be a collision if we do not brake. We forecast pedestrians analogously but model them as a point with velocity and acceleration. This works well because the movement patterns of pedestrians in CARLA are simple.} A collision infraction is detected if there is an intersection of the ego vehicle bounding box at future time step t with the bounding box of another traffic participant at future time step t. \red{A common failure of the action repeat forecast mechanism described above is that it does not anticipate that fast cars approaching the expert from behind will eventually slow down before colliding. To avoid false positives, we do not consider rear-end collisions by only using the front half of the vehicle as its bounding box.} For the longitudinal controller, we set $K_p=5.0, K_i=0.5, K_d=1.0$ and for the lateral controller, we set $K_p=1.25, K_i=0.75, K_d=0.3$. Both controllers use a buffer of size 40 to approximate the integral term as a running average. An example of the expert performing an unprotected left turn can be seen in \figref{fig:expert}.

\subsection{Longest6 Benchmark}
\label{sec:benchmark}

The CARLA simulator provides an official evaluation leaderboard consisting of 100 secret routes. However, teams using this platform are restricted to only 200 hours of evaluation time per month. A single evaluation takes over 100 hours, making the official leaderboard unsuitable for ablation studies or obtaining detailed statistics involving multiple evaluations of each model. Therefore, we propose the Longest6 Benchmark, which shares several similarities to the official leaderboard, but can be used for evaluation on local resources without computational budget restrictions.

The CARLA leaderboard repository provides a set of 76 routes as a starting point for training and evaluating agents. These routes were originally released along with the 2019 CARLA challenge. They span 6 towns and each of them is defined by a sequence of waypoints. However, there is a large imbalance in the number of routes per town, \eg Town03 and Town04 have 20 routes each, but Town02 has only 6 routes. To balance the Longest6 driving benchmark across all available towns, we choose the 6 longest routes per town from the set of 76 routes. This results in 36 routes with an average route length of 1.5km, which is similar to the average route length of the official leaderboard (1.7km). We make three other design choices for the Longest6 benchmark, motivated by the official leaderboard. (1) During evaluation, we ensure a high density of dynamic agents by spawning vehicles at every possible spawn point permitted by the CARLA simulator. (2) Following \cite{Chitta2021ICCV}, each route has a unique environmental condition obtained by combining one of 6 weather conditions (Cloudy, Wet, MidRain, WetCloudy, HardRain, SoftRain) with one of 6 daylight conditions (Night, Twilight, Dawn, Morning, Noon, Sunset). (3) We include CARLA's adversarial scenarios in the evaluation, which are spawned at predefined positions along the route. Specifically, we include CARLA's scenarios 1, 3, 4, 7, 8, 9, 10 which are generated based on the NHTSA pre-crash typology\footnote{\url{https://leaderboard.carla.org/scenarios/}}. \red{Visualizations of the routes in the Longest6 benchmark are provided in the supplementary material.}

\subsection{Metrics}
\label{sec:metrics}

We report several metrics to provide a comprehensive understanding of the driving behavior of each agent.

\noindent (1) \textbf{Route Completion (RC)}: percentage of route distance completed, $R_i$ by the agent in route $i$, averaged across $N$ routes. However, if an agent drives outside the route lanes for a percentage of the route, then the RC is reduced by a multiplier (1- \% off route distance).
\begin{equation}
    \text{RC} = \frac{1}{N} \sum_{i}^{N} {R_i}
\label{eqn:route_completion}
\end{equation}

\noindent (2) \textbf{Infraction Score (IS)}: geometric series of infraction penalty coefficients, $p^j$ for every instance of infraction $j$ incurred by the agent during the route. Agents start with an ideal 1.0 base score, which is reduced by a penalty coefficient for every infraction.
\begin{equation}
    \text{IS} = \prod_j^{\text{Ped,Veh,Stat,Red}} (p^j)^{\text{\# infractions}^j}
\label{eqn:infraction_multiplier}
\end{equation}
The penalty coefficient for each infraction is pre-defined and set to 0.50 for collision with a pedestrian, 0.60 for collision with a vehicle, 0.65 for collision with static layout, and 0.7 for red light violations. The official CARLA leaderboard also mentions a penalty for stop sign violations. However, we observe that none of our submissions have any stop sign infractions. Hence, we omit this infraction from our analysis.
\BlankLine

\noindent (3) \textbf{Driving Score (DS)}: weighted average of the route completion with infraction multiplier $P_i$
\begin{equation}
    \text{DS} = \frac{1}{N} \sum_{i}^{N} {R_i P_i}
\end{equation}

\noindent (4) \textbf{Infractions per km}: the infractions considered are collisions with pedestrians, vehicles, and static elements, running a red light, off-road infractions, route deviations, timeouts, and vehicle blocked.
\red{We report the total number of infractions, normalized by the total number of km driven.}
\begin{equation}
    \text{\red{Infractions per km}} = \frac{\sum_{i}^{N} {\text{\# infractions}_i}}{\sum_{i}^{N} {k_i}}
\end{equation}
\red{where $k_i$ is the driven distance (in km) for route $i$. The Off-road infraction is slightly different. Instead of the total \textit{number} of infractions the sum of \textit{km driven off-road} is used. We multiply by $100$ because this metric is a percentage.}

\begin{table*}[t]
\small
\centering
    \setlength{\tabcolsep}{3pt}
    \begin{tabular}{c| c c c | c c c c c c c c}
        \textbf{Method} & \textbf{DS} $\uparrow$ & \textbf{RC} $\uparrow$ & \textbf{IS} $\uparrow$ & \textbf{Ped} $\downarrow$ & \textbf{Veh} $\downarrow$ & \textbf{Stat} $\downarrow$ & \textbf{Red} $\downarrow$ & \textbf{OR} $\downarrow$ & \textbf{Dev} $\downarrow$ & \textbf{TO} $\downarrow$ & \textbf{Block} $\downarrow$ \\
        \hline
        WOR~\cite{Chen2021ICCVb} & \red{20.53} $\pm$ \red{3.12} & \red{48.47} $\pm$ \red{3.86} & \textbf{\red{0.56}} $\pm$ \red{0.03} & \red{0.18} & \red{1.05} & \red{0.37} & \red{1.28} & \red{0.47} & \red{0.88} & \red{0.08} & \red{0.20} \\
        Latent TransFuser (Ours) & \red{37.31} $\pm$ \red{5.35} & \textbf{\red{95.18}} $\pm$ \red{0.45} & \red{0.38} $\pm$ \red{0.05} & \red{\textbf{0.03}} & \red{3.66} & \red{0.18} & \red{0.13} & \red{\textbf{0.04}} & \red{\textbf{0.00}} & \red{0.12} & \red{\textbf{0.05}}\\
        \hline
        \red{LAV~\cite{Chen2022CVPR}} & \red{32.74} $\pm$ \red{1.45} & \red{70.36} $\pm$ \red{3.14} & \red{0.51} $\pm$ \red{0.02} & \red{0.16} & \red{\textbf{0.83}} & \red{0.15} & \red{0.96} & \red{0.42} & \red{0.06} & \red{0.12} & \red{0.45} \\
        Late Fusion (LF) & \red{22.47} $\pm$ \red{3.71} & \red{83.30} $\pm$ \red{3.04} & \red{0.27} $\pm$ \red{0.04} & \red{0.05} & \red{4.63} & \red{0.28} & \red{\textbf{0.11}} & \red{0.48} & \red{0.02} & \red{0.11} & \red{0.21} \\
        Geometric Fusion (GF) & \red{27.32} $\pm$ \red{0.80} & \red{91.13} $\pm$ \red{0.95} & \red{0.30} $\pm$ \red{0.01} & \red{0.06} & \red{4.64} & \red{0.17} & \red{0.13} & \red{0.48} & \textbf{\red{0.00}} & \textbf{\red{0.05}} & \red{0.11} \\
        TransFuser (Ours) & \textbf{\red{47.30}} $\pm$ \red{5.72} & \red{93.38} $\pm$ \red{1.20} & \red{0.50} $\pm$ \red{0.06} & \red{\textbf{0.03}} & \red{2.45} & \red{\textbf{0.07}} & \red{0.16} & \red{\textbf{0.04}} & \red{\textbf{0.00}} & \red{0.06} & \red{0.10} \\
        \hline
        \textit{Expert} & \textit{76.91 $\pm$ 2.23} & \textit{88.67 $\pm$ 0.56} & \textit{0.86 $\pm$ 0.03} & \textit{\red{0.02}} & \textit{\red{0.28}} & \textit{\red{0.01}} & \textit{\red{0.03}} & \textit{\red{0.00}} & \textit{\red{0.00}} & \textit{\red{0.08}} & \textit{\red{0.13}}\\
        \hline
    \end{tabular}
    \caption{\textbf{Longest6 Benchmark Results.} We compare our TransFuser model with several baselines in terms of driving performance and infractions incurred. We report the metrics for 3 evaluation runs of each model on the Longest6 evaluation setting. For the primary metrics (DS: Driving Score, RC: Route Completion, IS: Infraction Score) we show the mean and std. For the remaining infractions per km metrics (Ped: Collisions with pedestrians, Veh: Collisions with vehicles, Stat: Collisions with static layout, Red: Red light violation, OR: Off-road driving, Dev: Route deviation, TO: Timeout, Block: Vehicle Blocked) we show only the mean. TransFuser obtains the best DS \red{by a large margin}.}
    \label{tab:detailed_results}
    \vspace{0.0cm}
\end{table*}

\subsection{Baselines}
\label{sec:baselines}

We compare our TransFuser model to several baselines. (1) \textbf{WOR}~\cite{Chen2021ICCVb}: this is a multi-stage training approach that supervises the driving task with a Q function obtained using dynamic programming. It is the current state-of-the-art approach on the simpler NoCrash benchmark~\cite{Codevilla2019ICCV} for CARLA version 0.9.10. We use the author-provided pretrained model for evaluating this approach. (2) \textbf{Latent TransFuser}: to investigate the importance of the LiDAR input, we implement an auto-regressive waypoint prediction network that has the same architecture as the TransFuser but uses a fixed positional encoding image as input instead of the BEV LiDAR, as described in \secref{sec:ltf}. \red{(3) \textbf{LAV}~\cite{Chen2022CVPR}: this is a concurrent approach that performs sensor fusion via PointPainting~\cite{Vora2020CVPR}, which concatenates semantic class information extracted from the RGB image to the LiDAR point cloud, to train a privileged motion planner to predict trajectories of all nearby vehicles in the scene. This privileged planner is then distilled into a policy that drives from raw sensor inputs only. We use the checkpoint publicly released by the authors\footnote{\url{https://github.com/dotchen/LAV}} for our experiments. We note that this published checkpoint is not the exact same model as the one used for LAV's leaderboard entry.} (4) \textbf{Late Fusion}: we implement a version of our architecture where the image and the LiDAR features are extracted independently using the same encoders as TransFuser but without the transformers (similar to~\cite{Sobh2018NEURIPSW}). The features from each branch are then fused through element-wise summation and passed to the waypoint prediction network. (5) \textbf{Geometric Fusion}: we implement a multi-scale geometry-based fusion method, inspired by~\cite{Liang2018ECCV, Liang2019CVPR}, involving both image-to-LiDAR and LiDAR-to-image feature fusion. We unproject each 0.125m $\times$ 0.125m block in our LiDAR BEV representation into 3D space, resulting in a 3D cell. We randomly select 5 points from the LiDAR point cloud lying in this 3D cell and project them into the image space. Then, we aggregate the image features of these points via element-wise summation before passing them to a 3-layer MLP. The output of the MLP is then combined with the LiDAR BEV feature of the corresponding 0.125m $\times$ 0.125m block at multiple resolutions throughout the feature extractor. Similarly, for each image pixel, we aggregate information from the LiDAR BEV features at multiple resolutions. This baseline is equivalent to replacing the transformers in our architecture with projection-based feature fusion. We also report results for the expert used for generating our training data, which defines an upper bound for the performance on the Longest6 evaluation setting. We provide additional details regarding the baselines in the supplementary material. 

\subsection{Implementation Details}
\label{sec:impl}
 
We use 2 sensor modalities, the front-facing cameras and LiDAR point cloud converted to a BEV representation (\secref{sec:io_parameterization}), \ie, $S=2$. The camera inputs are concatenated to a single image and encoded using a RegNetY-3.2GF~\cite{Radosavovic2020CVPR} which is pretrained on ImageNet~\cite{Deng2009CVPR}. \red{We use pre-trained models from~\cite{Wightman2019TIMM}.} The LiDAR BEV representation is encoded using another RegNetY-3.2GF~\cite{Radosavovic2020CVPR} which is trained from scratch. Similar to \cite{Chen2019CORL}, we perform angular viewpoint augmentation for the training data, by randomly rotating the sensor inputs by $\pm$20$^\circ$ and adjusting the waypoint labels according to this rotation. We use 1 transformer per resolution and 4 attention heads for each transformer. We select $D_q, D_k, D_v$ from $\{72,216,576,1512\}$ for the 4 transformers corresponding to the feature embedding dimension $D_f$ at each resolution. We train the models with 4 RTX 2080Ti GPUs for 41 epochs, with an initial learning rate of $10^{-4}$ \red{and a batch size of $12$}. We reduce the learning rate by a factor of 10 after epoch 30 and 40. We evaluate the epochs 31, 33, 35, 37, 39 and 41 closed loop on the validation routes of \cite{Chitta2021ICCV} for one seed and pick the epoch with the highest driving score. For all models, we use the AdamW optimizer~\cite{Loshchilov2019ICLR}, which is a variant of Adam. Weight decay is set to $0.01$, and Adam beta values to the PyTorch defaults of $0.9$ and $0.999$.

\subsection{Longest6 Benchmark Results}
\label{sec:results_internal}

We begin with an analysis of driving performance on CARLA on the new Longest6 evaluation setting (\tabref{tab:detailed_results}). For each experimental result, the evaluation is repeated 3 times to account for the non-determinism of the CARLA simulator. Furthermore, imitation based methods typically show variation in performance when there is a change in the random initialization and data sampling due to the training seed~\cite{Behl2020IROS,Prakash2021CVPR}. To account for the variance observed between different training runs, we use 3 different random seeds for each method, and report the metrics for an ensemble of these 3 training runs.

\boldparagraph{Latent TransFuser as a Strong Baseline} In our first experiment, we examine the performance of image-based methods. From the results in \tabref{tab:detailed_results}, we observe that WOR performs poorly on the Longest6 evaluation setting. In particular, we observe that WOR suffers from a poor RC with a much larger number of route deviations (Dev) than the remaining methods. On the other hand, we find that Latent Transfuser obtains the best RC in \tabref{tab:detailed_results}, \red{with zero route deviations}. This is likely because Latent TransFuser uses our inverse dynamics model (PID controller) for low-level control and represents goal locations in the same BEV coordinate space in which waypoints are predicted. In contrast, WOR uses coarse navigational commands (\eg follow lane, turn left/right, change lane) to inform the model regarding driver intentions, and chooses an output action from a discrete action space. This result indicates that the TransFuser architecture involving auto-regressive waypoint prediction a strong baseline for the end-to-end driving task, even in the absence of a LiDAR sensor. 

\begin{figure*}[t]
    \centering
    \includegraphics[width=\textwidth]{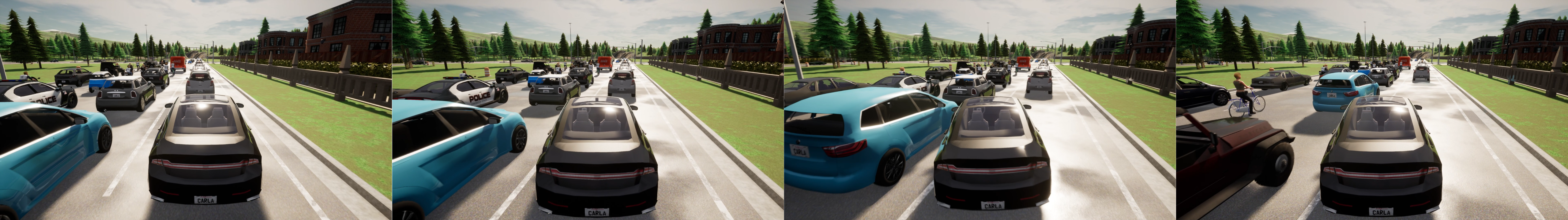}
    \includegraphics[width=\textwidth]{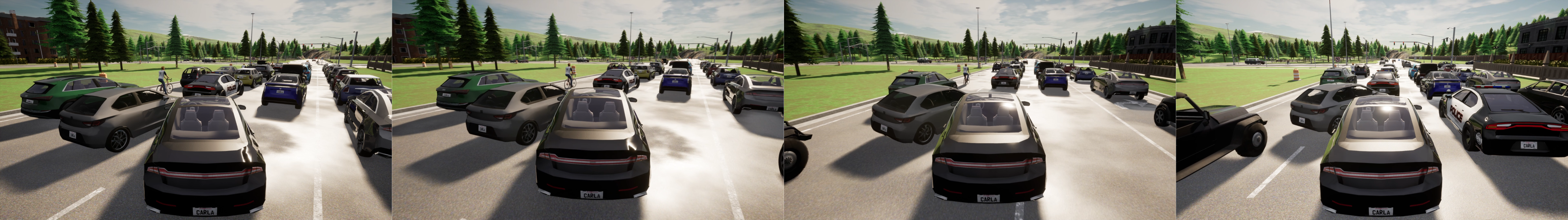}
    \caption{\red{\textbf{Lane Change Failures.} TransFuser fails at lane changes in dense traffic incurring a high number of consecutive collisions in routes where these situations occur. Two examples are shown in the top and bottom rows. Time goes forward from left to right.}}
    \vspace{-0.0cm}
    \label{fig:lane_change_collision}
    \vspace{-0.0cm}
\end{figure*}

\boldparagraph{Sensor Fusion Methods} The goal of this experiment is to determine the impact of the LiDAR modality on the driving performance and compare different fusion methods. \red{For this, we compare TransFuser to three baselines, LAV, Late Fusion (LF), Geometric Fusion (GF)}. \red{LAV performs worse than TransFuser in terms of DS. The main difference arises from the 23\% lower RC. Potential reasons could be worse steering as indicated by the higher off-road infractions, or false positives in the modular components which might be the reason for the higher blocked infraction. While LAV obtains a similar IS to TransFuser, upon probing further, we notice that it is only better in terms of avoiding collisions with vehicles and TransFuser performs better with respect to all other infractions. We note that in the Longest6 benchmark, there are a few routes where the vehicle is required to drive in dense traffic on multi-lane highways. TransFuser fails at lane merging in these situations, incurring a large amount of vehicle collisions ($>$20), strongly affecting its vehicle collision metric.} Surprisingly, we observe that LF and GF perform worse than the image-only Latent TransFuser baseline (\tabref{tab:detailed_results}). The multi-scale geometry-based fusion encoder of GF gives some improvements when compared to LF, however, both LF and GF suffer from a poor IS. We hypothesize that this occurs because they do not incorporate global contextual reasoning which is necessary to safely navigate the intersections, and focus primarily on navigation to the goal at all costs while ignoring obstacles, which leads to several infractions. In contrast, our TransFuser model obtains an absolute improvement \red{19.98\%} in terms of DS when compared to GF. It also achieves an \red{48.59\%} reduction compared to LF and \red{47.64\%} reduction compared to GF in collisions per kilometer \red{(Ped+Veh+Stat), and an absolute improvement of over 0.2 in its IS}. These results indicate that attention is effective in incorporating the global context of the 3D scene, which allows for safer driving. 

\begin{table}[t]
\small
    \setlength{\tabcolsep}{6pt}
    \centering
    \begin{tabular}{c| c c}
        \textbf{Method} & \textbf{Single Model} & \textbf{Ensemble (3)} \\
        \hline
        Late Fusion (LF) & 23.5 & 46.7 \\
        Geometric Fusion (GF) & 43.5 & 69.1 \\
        TransFuser (Ours) & 27.6 & 59.6 \\
        \hline
    \end{tabular}
    \caption{\textbf{Runtime.} We show the runtime per frame in ms for each method averaged over all timesteps in a single evaluation route. We measure runtimes for both a single model and an ensemble of three models. A single TransFuser model runs in real-time on an RTX 3090 GPU.}
    \label{tab:runtime}
    \vspace{-0.0cm}
\end{table}

\boldparagraph{Limitations} We observe that all methods \red{with a high RC} struggle with vehicle collisions (Veh). Avoiding collisions is very challenging in our evaluation setting due to the traffic density being set to the maximum allowed density in CARLA. In particular, TransFuser has around \red{9}$\times$ more vehicle collisions per kilometer than the expert. We observe that these collisions primarily occur during unprotected turns and lane changes \red{as is illustrated in \figref{fig:lane_change_collision}}.

\boldparagraph{Runtime} We measure the runtime of each method on a single RTX 3090 GPU by averaging over all time-steps of one evaluation route. The runtime considered includes sensor data pre-processing, model inference and PID control. The results are shown in \tabref{tab:runtime}. We observe that the transformers in our architecture increase the runtime relative to the LF baseline by 17\% for a single model and 28\% for an ensemble of three models. However, a single TransFuser model can still be executed in real-time on this hardware. The GF baseline is slower than TransFuser despite its simpler fusion mechanism due to the extra time taken finding correspondences between the image and LiDAR tokens.

\subsection{Leaderboard Results}
\label{sec:results_leaderboard}

We submit the models from our study to the CARLA autonomous driving leaderboard which contains a secret set of 100 evaluation routes and report the results in \tabref{tab:leaderboard}. Among the models that do not use LiDAR inputs, Latent TransFuser achieves the best performance. \red{It obtains a DS of 45.20, which is nearly 10 points better than the next best image-based method, GRIAD~\cite{Chekroun2021ARXIV}}. GRIAD builds on top of the Reinforcement Learning (RL) method presented in~\cite{Toromanoff2020CVPR}. In this approach, an encoder is first trained to predict both the 2D semantics and specific affordances such as the scene traffic light state, and the relative position and orientation between the vehicle and lane. The encoder is then frozen and used to train a value function-based RL method using a replay buffer that is partially filled with expert demonstrations. GRIAD requires 45M samples from the CARLA simulator for training ~\cite{Chekroun2021ARXIV} whereas our training dataset has only 228k frames (200$\times$ less than GRIAD). 

\begin{table}[t]
\small
	\centering
	\setlength{\tabcolsep}{5pt}
	\begin{tabular}{c | c | c c c}
	    \textbf{{Method}} & \textbf{{LiDAR?}} & \textbf{DS} $\uparrow$ & \textbf{RC} $\uparrow$ & \textbf{IS} $\uparrow$ \\
	    \hline
	    NEAT~\cite{Chitta2021ICCV} & - & 21.83 & 41.71 & 0.65 \\
	    MaRLn~\cite{Toromanoff2020CVPR} & - & 24.98 & 46.97 & 0.52 \\
	    WOR~\cite{Chen2021ICCVb} & - & 31.37 & 57.65 & 0.56 \\
	    GRIAD~\cite{Chekroun2021ARXIV} & - & 36.79 & 61.86 & 0.60 \\
		Latent TransFuser (Ours) & - & \red{45.20} & \red{66.31} & \textbf{\red{0.72}} \\
		\hline
 		Late Fusion (LF) & $\checkmark$ & \red{26.07} & \red{64.67} & \red{0.47} \\
 		Geometric Fusion (GF) & $\checkmark$  & \red{41.70} & \red{87.85} & \red{0.47} \\
		TransFuser (Ours) & $\checkmark$  & \red{61.18} & \red{86.69} & {\red{0.71}} \\
		LAV*~\cite{Chen2022CVPR} & $\checkmark$  & \textbf{\red{61.85}} & \textbf{\red{94.46}} & \red{0.64} \\
		\hline
	\end{tabular}
	\caption{\red{\textbf{CARLA Leaderboard Evaluation.} We report the DS, RC, and IS over the 100 secret routes of the official evaluation server. Latent TransFuser and TransFuser improve the IS by a large margin in comparison to existing methods. *The LAV leaderboard entry uses an updated model different from the public checkpoint in \tabref{tab:detailed_results}.}}
	\label{tab:leaderboard}
	\vspace{0.0cm}
\end{table}

\begin{figure*}[t]
    \centering
    \includegraphics[width=\textwidth]{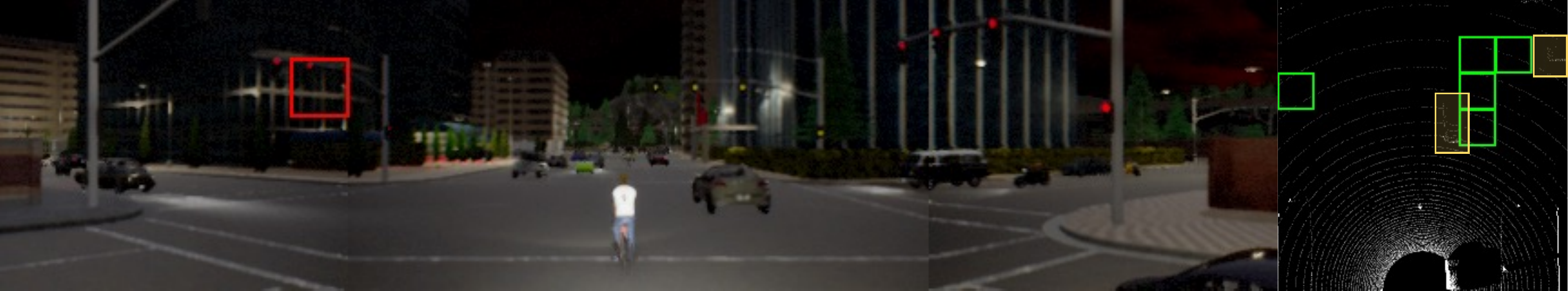}
    \includegraphics[width=\textwidth]{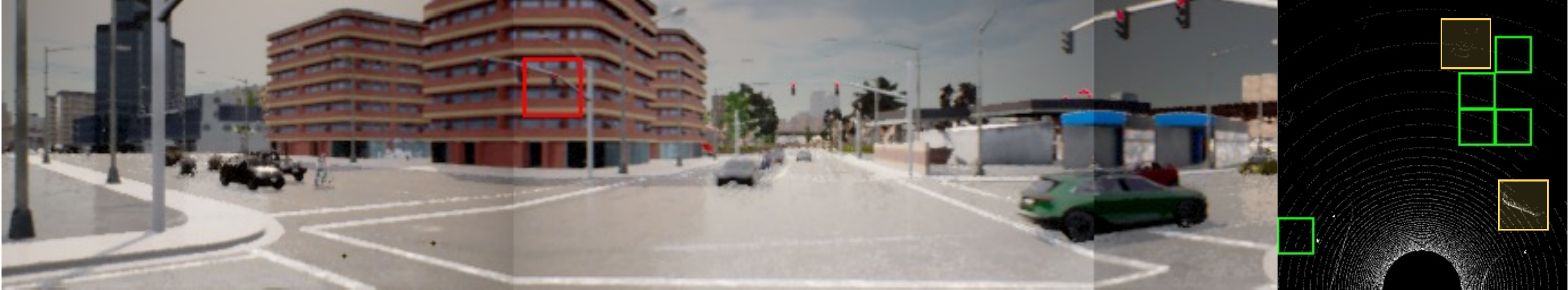}
    \includegraphics[width=\textwidth]{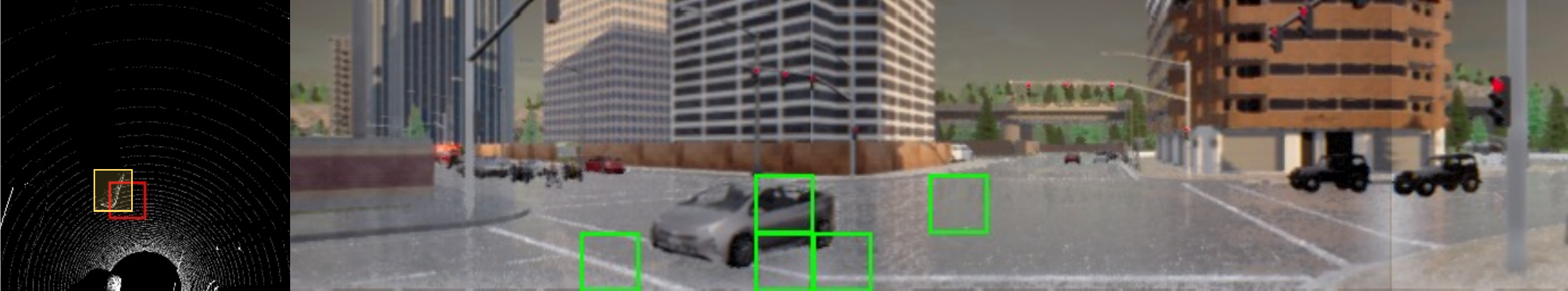}
    \includegraphics[width=\textwidth]{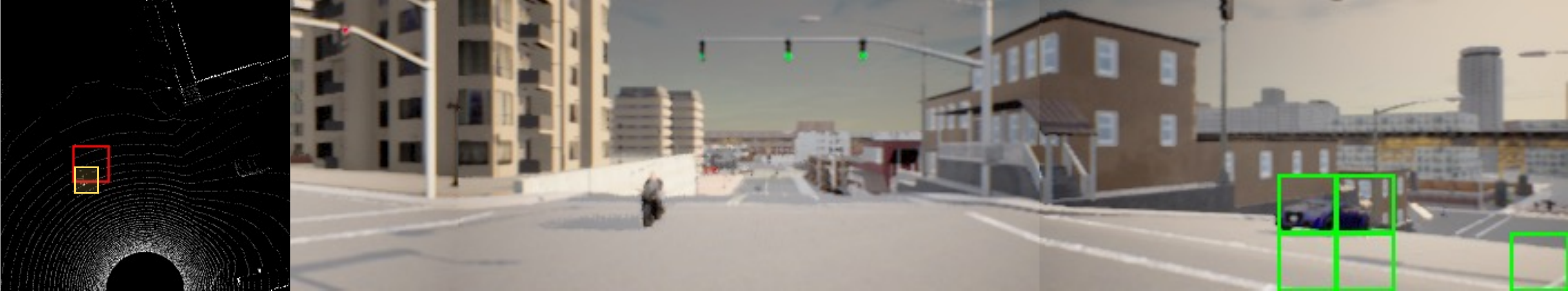}
    \caption{\textbf{Attention Maps.} For the {\color{red}{red}} query token, we show the top-5 attended tokens in {\color{darkgreen}{green}} and highlight the presence of vehicles in the LiDAR point cloud in {\color{darkyellow}{yellow}}. Top 2 rows: image to LiDAR, bottom 2 rows: LiDAR to image. TransFuser attends to areas near vehicles and traffic lights at intersections.}
    \vspace{-0.0cm}
    \label{fig:attn_map}
    \vspace{-0.0cm}
\end{figure*}

\red{For the LiDAR-based baselines, GF performs better than LF, similar to our findings on Longest6. Incorporating global attention via TransFuser leads to further improvements with a state-of-the-art IS. While LAV performs similarly to TransFuser, it is only marginally better in terms of DS (+0.67), which is likely within the evaluation variance. To obtain this marginal improvement, LAV adopts multi-stage training with several pretrained modular components and a teacher-student distillation framework. In contrast, TransFuser achieves state-of-the-art results with a straightforward single-stage IL training procedure.} For further improvements, the training procedure of TransFuser can potentially be combined with techniques used in previous work on autonomous driving such as Active Learning~\cite{Chitta2018ARXIV,Chitta2019ARXIV,Haussmann2020IV}, DAgger~\cite{Ross2011AISTATS,Prakash2020CVPR}, \red{adversarial simulation~\cite{Wang2021CVPRb,Rempe2021CVPR,Hanselmann2022ARXIV}}, RL-based fine-tuning~\cite{Ohn-Bar2020CVPR} and teacher-student distillation~\cite{Chen2019CORL, Chen2021ICCVb, Zhao2020CORL, Chen2022CVPR}.

\begin{table}[t]
\small
    \centering
    \setlength{\tabcolsep}{3pt}
    \begin{tabular}{c | c c | c c | c c | c c}
        \textbf{Head} & \multicolumn{2}{c|}{\textbf{T1}} & \multicolumn{2}{c|}{\textbf{T2}} & \multicolumn{2}{c|}{\textbf{T3}} & \multicolumn{2}{c}{\textbf{T4}} \\
        \hline
        & I$_t$ & L$_t$ & I$_t$ & L$_t$ & I$_t$ & L$_t$ & I$_t$ & L$_t$ \\
        \hline
        1 & 100.00 & 0.00 & 99.83 & 0.00 & 44.55 & 98.69 & 77.79 & 89.97 \\
        2 & 100.00 & 0.00 & 99.99 & 0.00 & 07.58 & 98.98 & 80.05 & 95.91 \\
        3 & 100.00 & 0.00 & 39.71 & 0.00 & 27.73 & 98.09 & 90.08 & 99.98 \\
        4 & 99.99 & 1.45 & 99.99 & 0.26 & 99.99 & 99.98 & 80.13 & 99.47 \\
        \hline
    \end{tabular}
    \caption{\textbf{Cross-Modal Attention Statistics.} We report the \% of tokens (I$_t$: Image tokens, L$_t$: LiDAR tokens) for which at least 1 of the top-5 attended tokens belongs to the other modality for each head of the four transformers: T1, T2, T3, T4.}
    \label{tab:attention_statistics}
\end{table}

\begin{table*}[t]
\small
	\centering
	\setlength{\tabcolsep}{6pt}
	\begin{tabular}{c | c | c | c c c | c c c }
	    \multirow{2}{*}{\textbf{Method}} & \multirow{2}{*}{\textbf{Ensemble?}} & \multirow{2}{*}{\textbf{Safety Heuristic}} & \multicolumn{3}{c|}{\textbf{Longest6}} & \multicolumn{3}{c}{\textbf{Leaderboard}}\\
	    & & & {DS} $\uparrow$ & {RC} $\uparrow$ & {IS} $\uparrow$ & {DS} $\uparrow$ & {RC} $\uparrow$ & {IS} $\uparrow$ \\
	    \hline
		\multirow{3}{*}{Latent TransFuser} & - & Global & \textbf{\red{50.00} $\pm$ \red{1.13}} & \red{90.38} $\pm$ \red{3.32} & \textbf{\red{0.56} $\pm$ \red{0.02}} & \red{47.05} & \red{71.66} & \red{0.72} \\ %
		& - & Creep Only & \red{42.19} $\pm$ \red{5.49} & \red{94.84} $\pm$ \red{1.40} & \red{0.44} $\pm$ \red{0.06} & \red{42.36} & \red{86.67} & \red{0.51} \\ %
		& $\checkmark$ & Creep Only & \red{36.18} $\pm$ \red{5.36} & \textbf{\red{95.34} $\pm$ \red{2.20}} & \red{0.37} $\pm$ \red{0.05} & \red{45.03} & \red{75.37} & \red{0.62} \\ %
		\hline
		\multirow{3}{*}{TransFuser} & - & Global & \red{49.49} $\pm$ \red{8.63} & \red{90.67} $\pm$ \red{4.78} & \red{0.55} $\pm$ \red{0.09} & \red{41.93} & \red{58.55} & \red{\textbf{0.77}} \\
		& - & Creep Only & \red{42.51} $\pm$ \red{2.49} & \red{91.01} $\pm$ \red{0.83} & \red{0.46} $\pm$ \red{0.02} & \red{50.57} & \red{73.84} & \red{0.68} \\ %
		& $\checkmark$ & Creep Only & \red{47.30} $\pm$ \red{5.72} & \red{93.38} $\pm$ \red{1.20} & \red{0.50} $\pm$ \red{0.06} & \red{\textbf{61.18}} & \red{\textbf{86.69}} & \red{0.71} \\ %
		\hline
	\end{tabular}
	\caption{\red{\textbf{Impact of Global Safety Heuristic.} The heuristic leads to consistent minor improvements for Latent TransFuser. For TransFuser, though the heuristic improves the Longest6 scores, it reduces the performance on the CARLA leaderboard.}} 
	\label{tab:safety}
	\vspace{0.0cm}
\end{table*}

\subsection{Attention Statistics and Visualizations}
\label{sec:visualizations}

Our architecture (\figref{fig:model}) consists of 4 transformers with 4 attention layers and 4 attention heads in each transformer. In this section, we visualize the attention maps from the final attention layer for each head for each transformer. The transformer takes in 110 image feature tokens and 64 LiDAR feature tokens as input where each token corresponds to a $32 \times 32$ patch in the input modality. We consider intersection scenarios from Town03 and Town05 and examine the top-5 attention weights for the 66 tokens in the 2$^{nd}$, 3$^{rd}$ and 4$^{th}$ rows of the image feature map and the 24 tokens in the 4$^{th}$, 5$^{th}$ and 6$^{th}$ rows of the LiDAR feature map. We select these tokens since they correspond to the intersection region in the input modality and contain traffic lights and vehicles.

We compute statistics on cross-modal attention for image and LiDAR feature tokens. Specifically, we report the \% of tokens for which at least one of the top-5 attended tokens belong to the other modality for each head of the 4 transformers (T1, T2, T3, T4) in~\tabref{tab:attention_statistics}. We observe that the earlier transformers have negligible LiDAR to image attention compared to later transformers in which nearly all the LiDAR tokens aggregate information from the image features. Furthermore, different heads of each transformer also show distinctive behavior, \eg head 3 of T2 has significantly less cross-attention for image tokens than other heads, head 2 of T3 has very little cross-attention whereas head 4 has significantly higher cross-attention for image tokens compared to other heads. Overall, T4 exhibits extensive cross-attention for both image and LiDAR tokens, which indicates that TransFuser is effective in aggregating information between the two modalities.

We show four cross-attention visualizations for T4 in~\figref{fig:attn_map}. We observe a common trend in attention maps: TransFuser attends to areas near vehicles and traffic lights at intersections, albeit at a slightly different location in the image and LiDAR feature maps. Additional visualizations for all the transformers are provided in the supplementary.

\subsection{Global Safety Heuristic}
\label{sec:safety}
\red{The primary motivation of the safety heuristic described in \secref{sec:control} is to prevent collisions during the applied creeping behavior. Therefore, in Tables \ref{tab:detailed_results} and \ref{tab:leaderboard}, we apply the safety heuristic during creeping only. However, rule-based fallback systems have been shown to improve the safety of IL models~\cite{Vitelli2021ARXIV}. In this experiment, we investigate the impact of applying the safety heuristic described in \secref{sec:control} globally, \ie during both creeping and regular driving. We show results on both the Longest6 benchmark and the CARLA leaderboard. To reduce the computational overhead of this analysis, we evaluate a single model instead of an ensemble of 3 different training runs, which were used in Tables \ref{tab:detailed_results} and \ref{tab:leaderboard}. However, for clarity, we also report the scores of the ensemble. Additionally, the results shown for Latent TransFuser are from preliminary experiments where we include a LiDAR sensor and use the LiDAR-based safety heuristic instead of the CenterNet based intersection check described in \secref{sec:ltf}, leading to minor differences when compared to the numbers from Tables \ref{tab:detailed_results} and \ref{tab:leaderboard}.

The results are shown in \tabref{tab:safety}. For Latent TransFuser, we observe that the global safety heuristic improves the DS by 8 points on the Longest6 benchmark and 5 points on the leaderboard compared to it being applied during creeping only. In particular, this is due to a large improvement in the IS (\eg from 0.51 to 0.72 on the leaderboard). Interestingly, we observe a different trend for TransFuser. The global safety heuristic improves the DS by 7 points on the Longest6 benchmark where we tuned the hyper-parameters (\ie, size of the rectangular safety box). However, it leads to a drop of nearly 10 points in DS on the leaderboard. The global safety heuristic leads to reduced route completion for both methods on the CARLA leaderboard. This indicates that the heuristic works well on observed maps, but does not generalize to unknown and potentially unseen road layouts. For TransFuser, which already had a higher infraction score than Latent TransFuser, the improvement in IS does not make up for the reduction in RC leading to an overall reduction in DS through the safety heuristic.

When comparing the results of a single model and the corresponding ensemble, we find that ensembling improves the DS for both Latent TransFuser and TransFuser on the leaderboard, in particular for TransFuser which improves by more than 10 points. On the Longest6 routes, the ensembling has a lower impact of 5 points for the TransFuser and even a reduction in performance for Latent TransFuser. The single models reported in \tabref{tab:safety} are the first of three training seeds (\tabref{tab:trainseedvar}). Ensembling might have a larger positive impact on TransFuser because the models had a larger training variance, which we discuss in the following.}

\begin{table}[t]
\small
    \setlength{\tabcolsep}{6pt}
    \centering
    \begin{tabular}{c| c | c c c c }
        \textbf{Training} & \textbf{Eval} & \textbf{LTF} & \textbf{LF} & \textbf{GF} & \textbf{TF} \\
        \hline
        \multirow{4}{*}{1} & 1 & 48.82 & 31.94 & \red{46.13} & 59.45 \\
        & 2 & 50.12 & 33.29 & \red{43.62} & 44.15 \\
        & 3 & 51.07 & 34.79 & \red{39.42} & 44.87 \\
        & \red{avg.} & \red{50.00} & \red{33.34} & \red{43.06} & \red{49.49} \\
        \hline
        \multirow{4}{*}{2} & 1 & \red{44.53} & 37.05 & \red{36.64} & 51.50 \\
        & 2 & \red{54.35} & 36.79 & \red{39.40} & 50.91 \\
        & 3 & \red{52.57} & 47.98 & \red{34.93} & 52.35 \\
        & \red{avg.} & \red{50.48} & \red{40.60} & \red{36.99} & \red{51.59} \\
        \hline
        \multirow{4}{*}{3} & 1 & \red{51.15} & 34.90 & \red{49.77} & 59.76 \\
        & 2 & \red{48.10} & 32.47 & \red{45.64} & 57.39 \\
        & 3 & \red{49.80} & 44.98 & \red{52.47} & 52.88 \\
        & \red{avg.} & \red{49.68} & \red{37.45} & \red{49.30} & \red{56.68} \\
        \hline
    \end{tabular}
    \caption{\textbf{Training and evaluation variance.} We show the DS of each evaluation on the Longest6 benchmark. We train each baseline 3 times, and perform 3 evaluation runs of each individual trained model. LTF: Latent TransFuser, LF: Late Fusion, GF: Geometric Fusion, TF: TransFuser. All models exhibit large variance in scores.}
    \label{tab:trainseedvar}
    \vspace{-0.0cm}
\end{table}

\boldparagraph{Training Seed Variance} We show the impact of training and evaluation seed variance in \tabref{tab:trainseedvar}. We train each baseline 3 times and evaluate each model 3 times on the Longest6 routes \red{with the global safety heuristic enabled}. We observe that the best achieved score can differ from the worst score by 10-15 points for an individual model, leading to extremely large variance. In particular, for the first seed of TransFuser, the DS ranges from 44.15 to 59.45. For Geometric Fusion, \red{the average DS differs by 12 points between the worst and best training seed.} This amount of variance is problematic when trying to analyze or compare different approaches. We would like to emphasize that the variance reported in \tabref{tab:trainseedvar} comes from two factors. The training variance between different seeds results from different network initializations, data sampling and data augmentations during optimization. The evaluation variance is a result of the variation in the traffic manager, physics and sensor noise of CARLA 0.9.10. Based on the results observed, the randomness in evaluation is the primary cause of variance, in addition to secondary training seed variance, but both factors are considerable. The existing practice for state-of-the-art methods on CARLA is to report only the evaluation variance by running multiple evaluations of a single training seed. \red{This may lead to premature conclusions (\eg when considering only the three evaluations of the first training seed, Latent TransFuser outperforms TransFuser).} We argue (given these findings) that future studies should report results by varying the training seed for both the baselines and proposed new methods, in addition to the results of the best seed or ensemble.

\subsection{Ablation Studies}
\label{sec:ablation}

We now analyze several design choices for TransFuser in a series of ablation studies on the Longest6 benchmark. \red{Since the global safety heuristic leads to consistent and significant improvements for TransFuser on the Longest6 routes (\tabref{tab:safety}), we use this setting for the ablation studies.} The evaluation is repeated 3 times for each experiment, however, we use a single training run for these results instead of an ensemble of 3 different training runs as in \tabref{tab:detailed_results}. \red{To further reduce the computational overhead, we always evaluate epoch 31, as we have observed in preliminary experiments that it is usually close to the best epoch in performance.} \red{For the default configuration, we have 3 available training runs. We report the best and worst training seed to account for the randomness due to training. Ablations lying within this interval likely do not have a large impact.}

\begin{figure}
    \begin{subfigure}[b]{0.48\textwidth}
        \centering
        \includegraphics[width=\textwidth]{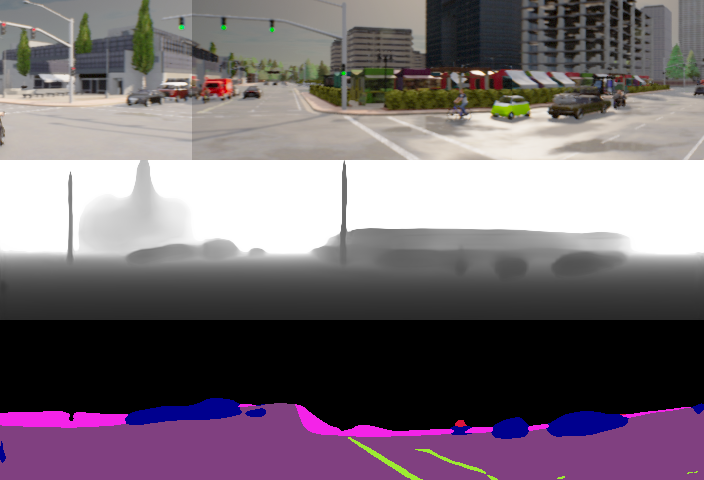}
        \caption{Top to bottom: image input, predicted depth, predicted semantics (legend: {\color{color_unlabled}none}, {\color{color_road}road}, {\color{color_roadline}lane}, {\color{color_sidewalk}sidewalk}, {\color{color_vehicle}vehicle}, {\color{color_person}person}).}
        \label{fig:qual1}
    \end{subfigure}
    \hfill
    \vspace{0.15cm}
    \begin{subfigure}[b]{0.48\textwidth}
        \centering
        \includegraphics[width=\textwidth]{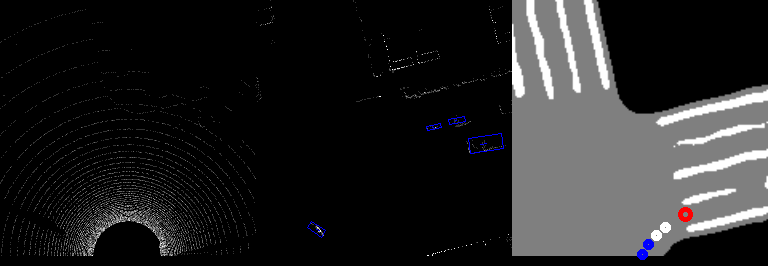}
        \caption{Left to right: LiDAR ground plane channel, bounding box predictions overlaid on LiDAR obstacle channel (points above ground plane), HD map prediction.}
        \label{fig:qual2}
    \end{subfigure}
    \vspace{0.0cm}
    \caption{\label{fig:qual}\textbf{Visualization of Auxiliary Tasks.} We visualize the inputs and outputs of both the image branch and LiDAR branch for the same driving scene. Further, the input target point is visualized as a red circle and predicted waypoints as blue and white circles on the HD map prediction. The first two waypoints (which are used to obtain the steering angle for our PID controller) are shown as blue circles, and the remaining two waypoints as white circles.}
\end{figure}

\boldparagraph{Auxiliary Tasks} As described in \secref{sec:loss}, we consider 4 auxiliary tasks in this work. In \tabref{tab:ablation_aux} we show the performance of TransFuser when all these auxiliary losses are removed, as well as the impact of removing each loss independently. We observe that with no auxiliary tasks, there is a steep drop in RC from 92.28 to 78.17. \red{Removing only a single auxiliary task does not have a large impact. All results lie within the range between the best and worst seed of the default configuration in terms of driving score.} In \figref{fig:qual}, we visualize the predictions made by TransFuser when trained with all 4 auxiliary losses.

\begin{table}[t]
\small
    \setlength{\tabcolsep}{8pt}
    \centering
    \begin{tabular}{c | c c c }
        \textbf{Auxiliary Losses} & \textbf{DS} $\uparrow$ & \textbf{RC} $\uparrow$ & \textbf{IS} $\uparrow$ \\
        \hline
        None & 44.29 & 78.17 & 0.58 \\
        \hline
        No Depth & \red{56.23} & \red{91.80} & 0.61\\
        No Semantics & 53.76 & 88.40 & 0.61\\
        No HD Map & 50.96 & 89.52&0.58 \\
        No Vehicle Detection & 53.43 & 88.49 & 0.60\\
        \hline
        \red{All Losses (Worst Seed)} & \red{49.49} & \red{90.67} & \red{0.55} \\
        All Losses \red{(Best Seed)} & \textbf{56.68} &  \textbf{92.28} &  \textbf{0.62}\\
        \hline
    \end{tabular}
    \caption{\textbf{Auxiliary Tasks.} The results shown are the mean over 3 evaluations on Longest6. \red{Training without auxiliary losses leads to a significant reduction in RC and DS.}}
    \label{tab:ablation_aux}
    \vspace{-0.0cm}
\end{table}

\begin{table}[t]
\small
    \setlength{\tabcolsep}{5pt}
    \centering
    \begin{tabular}{c| c | c c c}
        \textbf{Parameter} & \textbf{Value} & \textbf{DS} $\uparrow$ & \textbf{RC} $\uparrow$ & \textbf{IS} $\uparrow$ \\
        \hline
        \multirow{2}{*}{\red{Fusion Direction}} & \red{LiDAR $\rightarrow$ Camera} & \red{46.36} & \red{87.46} & \red{0.55} \\
         & \red{Camera $\rightarrow$ LiDAR} & \red{47.99} & \red{86.24} & \red{0.57} \\
         \hline
        \multirow{3}{*}{Fusion Scales} & 1 & 49.35 & 84.47&0.57 \\
        & 2 & 53.52 &91.78 & 0.59\\
        & 3 & 48.77 & 85.33& 0.60\\
        \hline
        \multirow{3}{*}{Attention \red{L}ayers} & 2 & 53.49 & 90.65 & 0.60 \\
        & 6 & 56.24 & {92.56} & 0.61\\
        & 8 & \red{56.13} & \red{\textbf{92.61}} & \red{0.61}\\
        \hline
         \multirow{2}{*}{\red{Token Count}} & \red{11$\times$3 + 8$\times$8} & \red{45.63} & \red{90.32} & \red{0.51} \\
         & \red{22$\times$5 + 4$\times$4} & \red{49.14} & \red{87.10} & \red{0.56} \\
        \hline
         \red{Averaging} & \red{Attention Token} & \red{54.21} & \red{90.78} & \red{0.60} \\
        \hline
        \multirow{4}{*}{Backbones} & Res34-Res18 & 42.01 & 92.43 & 0.45\\
         & Reg0.8-Reg0.8 & 44.38 & 87.96& 0.50\\
         & Reg1.6-Reg1.6 & 47.80 & 91.62& 0.52\\
         & \red{NeXt-T-NeXt-T} & \red{48.27} & \red{92.30} & \red{0.53} \\ 
        \hline
        \multirow{2}{*}{\red{Default Config}} & \red{Worst Seed} & \red{49.49} & \red{90.67} & \red{0.55} \\
        & \red{Best Seed} & \textbf{56.68} & 92.28 & \textbf{0.62} \\
        \hline
    \end{tabular}
    \caption{\textbf{Architecture Ablations.} The results shown are the mean over 3 evaluations on Longest6. \red{The default configuration fuses in both directions. It uses} 4 fusion scales, 4 attention layers, \red{22$\times$5 + 8$\times$8 tokens, global average pooling}, and RegNetY-3.2GF backbones. \red{The encoder backbone has the highest impact on the final driving score.}}
    \label{tab:ablation_arch}
    \vspace{-0.0cm}
\end{table}

\boldparagraph{Architecture} In \tabref{tab:ablation_arch}, we analyze the impact of varying the TransFuser encoder architecture. \red{We study the importance of fusion in both directions by selectively removing the residual output connections from the fusion transformers to the convolutional backbones. Fusion for only the Camera $\rightarrow$ LiDAR or only the LiDAR $\rightarrow$ Camera direction gives a slightly lower performance that the default model with bi-directional fusion.} Removing the fusion mechanism in the early blocks of the encoder and performing feature fusion at only the deepest 1, 2 and 3 scales also leads to a small drop in performance. For the fusion transformers, we find that \red{2-8} attention layers give similar performance. \red{The default resolution of the image features is 22$\times$5 and the LiDAR features is 8$\times$8. We observe that using each of these 22$\times$5 + 8$\times$8 features as independent input tokens for the transformer leads to better results when compared to a fusing information across different image and LiDAR resolutions through a reduced image token count of 11$\times$3 or LiDAR token count of 4$\times$4. We also evaluate a version of TransFuser where the input to the MLP and GRU decoder from the final fusion transformer is obtained via a dedicated attention token instead of the default global average pooling. This is a standard idea for attention-based averaging of spatial features, similar to the CLS token of Vision Transformers~\cite{Dosovitskiy2021ICLR}. We find that the default architecture with global average pooling is simpler to implement and performs similarly in practice.} The most impactful architecture choice is the backbone architecture for both branches. The default configuration of RegNetY-3.2GF backbones outperforms \red{ConvNeXt-Tiny~\cite{Liu2022ARXIV}}, RegNetY-1.6GF and RegNetY-0.8GF based backbones. We also observe a large improvement over the use of a ResNet34 for the image branch and ResNet18 for the BEV branch, as in \cite{Prakash2021CVPR}, which leads to a model with lower network capacity.

\boldparagraph{Model Inputs} As shown in \tabref{tab:ablation_input}, increasing the LiDAR range or reducing the camera FOV from the default configuration leads to a reduced IS and a corresponding drop in DS. \red{Our method works with arbitrary grids as inputs. Therefore, it could benefit from orthogonal improvements in LiDAR encoding. However, we did not observe improvements by using a learned LiDAR encoder~\cite{Lang2019CVPR}, and hence stick with the simpler voxelization approach. This model was trained with batch size 10 due to its higher memory requirements.} Interestingly, despite being useful in our preliminary experiments, removing the rasterized goal location channel from the LiDAR branch, or removing the random rotation of sensor inputs by $\pm$20$^\circ$ used during data augmentation show only a small impact on the performance in the final configuration which is unlikely to be significant.

\boldparagraph{Inertia Problem} As we note in \secref{sec:control}, we add creeping to our controller to prevent the agent from being overly cautious. This type of behavior, called the inertia problem~\cite{Codevilla2019ICCV} is typically attributed to the spurious correlation that exists between input velocity and output acceleration in an IL dataset. Interestingly, though we do not use velocity as an input to our models, we observe that creeping in the controller increases the RC significantly while maintaining a similar IS (\tabref{tab:ablation_creeping}). This indicates that a factor besides the velocity input, such as an imbalance in training data distribution, may be a key contributing factor to the inertia problem. We also train a version of TransFuser where we provide the current velocity as input by projecting the scalar value into the same dimensions as the transformer positional embedding using a linear layer. This velocity embedding is combined with the learnable positional embedding through element-wise summation and fed into the transformer at all 4 stages of the backbone. Including the velocity input leads to a sharp drop in DS, which cannot be recovered through the creeping behavior.

\begin{table}[t]
\small
    \setlength{\tabcolsep}{5pt}
    \centering
    \begin{tabular}{c| c | c c c}
        \textbf{Parameter} & \textbf{Value} & \textbf{DS} $\uparrow$ & \textbf{RC} $\uparrow$ & \textbf{IS} $\uparrow$ \\
        \hline
        \multirow{2}{*}{LiDAR Range} & 64m $\times$ 32m & 49.08 & 91.10 & 0.54\\
        & 64m $\times$ 64m & 47.55 & 90.72 & 0.52\\
        \hline
        \red{LiDAR Encoder} & \red{PointPillars} & \red{50.83} & \red{91.56} & \red{0.55}\\
        \hline
        \multirow{2}{*}{Camera FOV} & 120$^\circ$ & 49.90 & 90.05& 0.56\\
        & 90$^\circ$ & 42.18 & 88.49& 0.51\\
        \hline
        No Rasterized Goal & - & 54.80 & 91.63 & 0.60\\
        No Rotation Aug & - & \textbf{56.85} & \textbf{92.73} & 0.61 \\
        \hline
        \multirow{2}{*}{\red{Default Config}} & \red{Worst Seed} & \red{49.49} & \red{90.67} & \red{0.55} \\
        & \red{Best Seed} & 56.68 & 92.28 & \textbf{0.62} \\
        \hline
    \end{tabular}
    \caption{\textbf{Model Input Ablations.} The results shown are the mean over 3 evaluations on Longest6. The default configuration \red{uses} a 32m $\times$ 32m LiDAR range and 132$^\circ$ camera FOV. \red{Camera FOV has the largest impact on the DS.}}
    \label{tab:ablation_input}
    \vspace{-0.0cm}
\end{table}

\begin{table}[t]
\small
    \setlength{\tabcolsep}{6pt}
    \centering
    \begin{tabular}{c| c | c c c }
        \textbf{Velocity Input?} & \textbf{Creeping?} & \textbf{DS} $\uparrow$ & \textbf{RC} $\uparrow$ & \textbf{IS} $\uparrow$ \\
        \hline
        \multirow{2}{*}{-} & - & 46.35 & 78.28 & 0.63\\
        & $\checkmark$ & \textbf{56.68} & \textbf{92.28} & 0.62 \\
        \hline
        \multirow{2}{*}{$\checkmark$} & - & 37.34 & 64.27 & \textbf{0.65}\\
        & $\checkmark$ & 45.35 & 86.22 & 0.52\\
        \hline
    \end{tabular}
    \caption{\textbf{Inertia Problem.} The results shown are the mean over 3 evaluations on Longest6. Creeping improves the RC in both the setting where we input the velocity to our encoder and our default configuration (no velocity input).}
    \label{tab:ablation_creeping}
    \vspace{-0.0cm}
\end{table}

\section{Discussion and Conclusions}

In this work, we demonstrate that IL policies based on existing sensor fusion methods suffer from high infraction rates in complex driving scenarios. To overcome this limitation, we present a novel multi-modal fusion transformer (TransFuser) for integrating representations of different modalities. TransFuser uses attention to capture the global 3D scene context and focuses on dynamic agents and traffic lights, resulting in state-of-the-art performance on CARLA with a significant reduction in infractions. Given that our method is simple, flexible and generic, it would be interesting to explore it further with additional sensors, \eg radar, or apply it to other embodied AI tasks.

We hope that the proposed benchmark with long routes and dense traffic will become a suitable option for the community to conduct ablation studies or obtain detailed statistics that are not feasible via the CARLA leaderboard.

Our study has several limitations. We have provided a simple solution to the inertia problem (creeping), but this deserves more study. Due to the sensor limits of the CARLA leaderboard, our sensor setup does not generate data from the rear of the vehicle, which is relevant in lane change situations. We only investigate single time step input data in this work. Processing temporal inputs is likely necessary to reduce vehicle collisions in intersections by enabling estimation of the acceleration and velocity of other traffic participants. We do not investigate the impact of latency on the final driving performance, which has been shown to be important for real-world applications \cite{Li2020ECCVb}. This is because the default configuration of the CARLA simulator waits for the agent to finish its computation before it resumes simulation of the world. Finally, all our experiments are only conducted in simulation. Real-world data is more diverse and can have more challenging noise. We make use of multiple high-quality labels that the CARLA simulator provides, such as dense depth maps. Real-world datasets often do not provide labels of such high quality and might not provide all the types of labels we have used in this work.

Progress on the CARLA leaderboard has been rapid, with the state-of-the-art scores increasing from the range of 20 to \red{60} in the short time period since the preliminary version of this work at CVPR 2021. As novel submissions to the leaderboard move towards alternatives to end-to-end IL that involve complex multi-stage training or RL-based objectives, we show that a simple IL training procedure with our proposed architecture is highly competitive. Future works should consider our Latent Transfuser as a standard baseline for image-only IL. Based on our analysis, we believe that overcoming the inertia problem in a principled manner and reducing both training and evaluation variance will be key challenges for IL-based driving. 

\section*{Acknowledgements}

This work was supported by the BMWi in the project KI Delta Learning (project number: 19A19013O) and the German Federal Ministry of Education and Research (BMBF): T\"ubingen AI Center, FKZ: 01IS18039B. Andreas Geiger was supported by the ERC Starting Grant LEGO-3D (850533) and the DFG EXC number 2064/1 - project number 390727645. The authors thank the International Max Planck Research School for Intelligent Systems (IMPRS-IS) for supporting Kashyap Chitta, \red{Bernhard Jaeger} and Katrin Renz. The authors also thank Pavan Teja Varigonda for his help with sampling the training routes and Niklas Hanselmann for proofreading.

\bibliographystyle{IEEEtran}
\bibliography{bibliography_long,bibliography,bibliography_custom}

\begin{IEEEbiography}[{\includegraphics[width=1in,height=1.25in,clip,keepaspectratio]{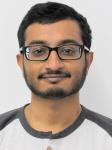}}]{Kashyap Chitta}
is a Ph.D. student with the Autonomous Vision Group led by Prof. Andreas Geiger, part of the Max Planck Institute for Intelligent Systems and University of T\"ubingen, Germany. He obtained his M.S. in Computer Vision in 2018 from Carnegie Mellon University under the supervision of Prof. Martial Hebert. During this time, he also worked as a Deep Learning Intern at NVIDIA on two occasions. 
\end{IEEEbiography}
\begin{IEEEbiography}[{\includegraphics[width=1in,height=1.25in,clip,keepaspectratio]{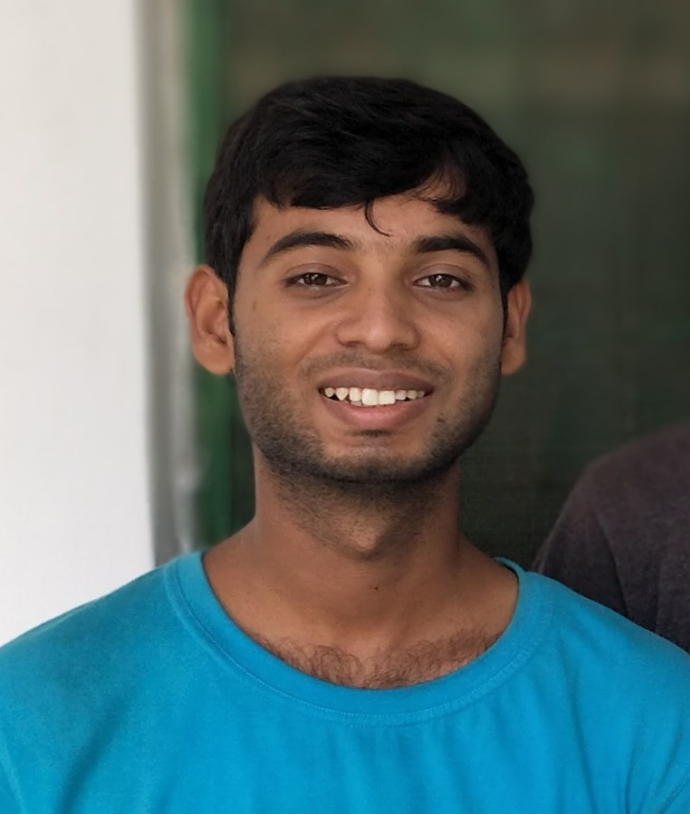}}]{Aditya Prakash}
is a Ph.D. student at the University of Illinois Urbana Champaign, advised by Prof. Saurabh Gupta. Prior to this, he worked as a research assistant at the Max Planck Institute for Intelligent Systems T\"ubingen as part of the Autonomous Vision Group led by Prof. Andreas Geiger. He obtained his bachelors at the Indian Institute of Technology Roorkee and has worked as an intern at Adobe Research, Indian Institute of Science and NAVER Clova AI Research.
\end{IEEEbiography}
\begin{IEEEbiography}
[{\includegraphics[width=1in,height=1.25in,clip,keepaspectratio]{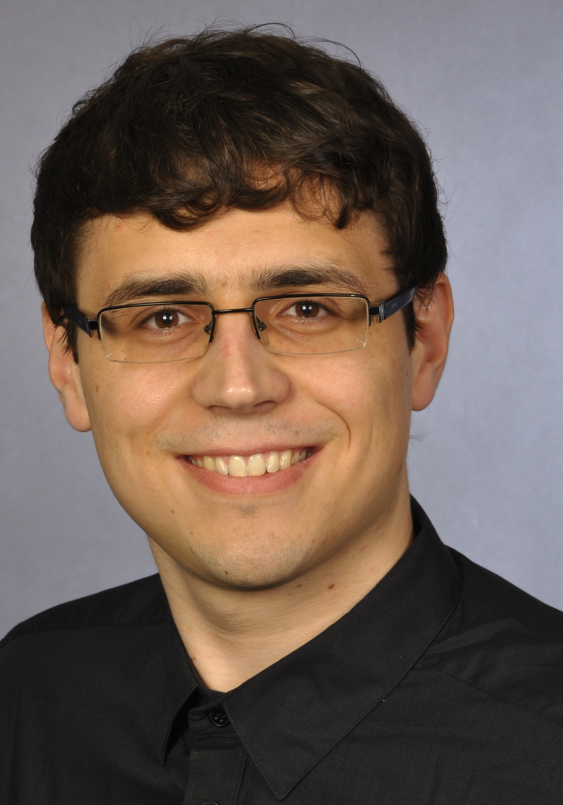}}]{Bernhard Jaeger}
\red{is a Ph.D. student with the Autonomous Vision Group led by Prof. Andreas Geiger, part of the International Max Planck Research School for Intelligent Systems and University of T\"ubingen, Germany.} He obtained his M.Sc. in Computer Science in 2021 from University of T\"ubingen and his B.Sc. degree in Informatics: Games Engineering from the Technical University of Munich in 2018. He worked for 1 year at FERCHAU GmbH as a software developer.
\end{IEEEbiography}
\begin{IEEEbiography}[{\includegraphics[width=1in,height=1.25in,clip,keepaspectratio]{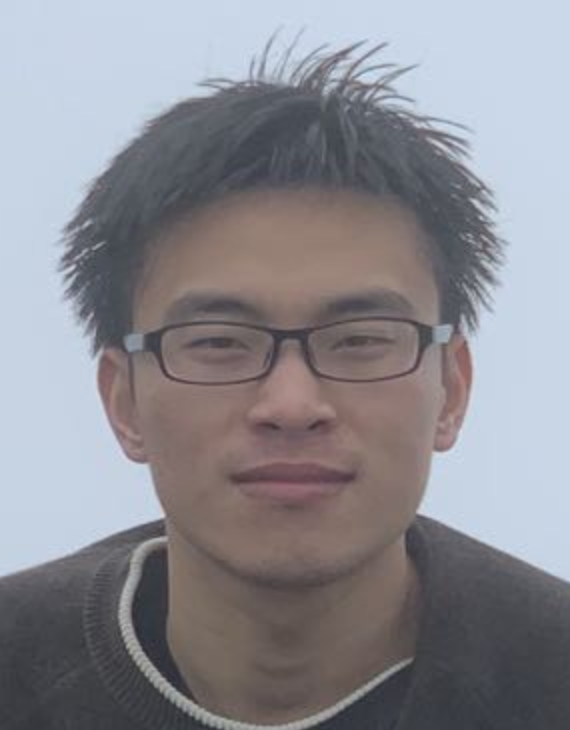}}]{Zehao Yu} is a Ph.D. student with the Autonomous Vision Group led by Prof. Andreas Geiger, at the University of T\"ubingen, Germany. He obtained his M.S. in Computer Science in 2021 from ShanghaiTech University under the supervision of Prof. Shenghua Gao and his B.Sc. degree in Software Engineering from Xiamen University. 
\end{IEEEbiography}
\begin{IEEEbiography}
[{\includegraphics[width=1in,height=1.25in,clip,keepaspectratio]{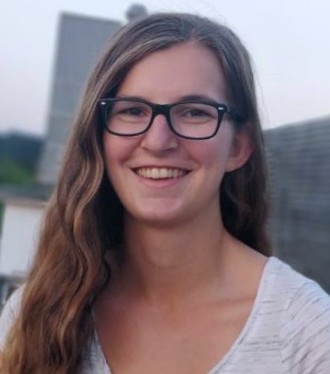}}]{Katrin Renz} is a Ph.D. student with the Autonomous Vision Group led by Prof. Andreas Geiger, part of the Max Planck Institute for Intelligent Systems and University of T\"ubingen, Germany. She obtained her masters in 2021 from the University of Heilbronn and has spent time as a visiting student in the Visual Geometry Group at the University of Oxford. 
\end{IEEEbiography}
\begin{IEEEbiography}
[{\includegraphics[width=1in,height=1.25in,clip,keepaspectratio]{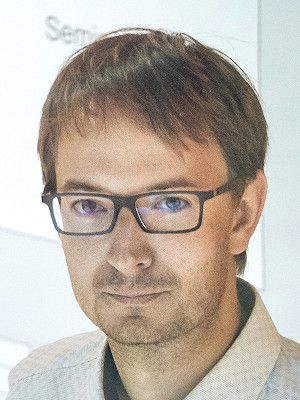}}]{Andreas Geiger} is a professor at the University of Tübingen. Prior to this, he was a visiting professor at ETH Zürich and a group leader at the Max Planck Institute for Intelligent Systems. He studied at KIT, EPFL and MIT, and received his PhD degree in 2013 from the Karlsruhe Institute of Technology (KIT). His research interests are at the intersection of computer vision, machine learning and robotics, with a particular focus on 3D scene perception, deep representation learning, generative models and sensori-motor control in the context of autonomous systems. He maintains the KITTI and KITTI-360 benchmarks.
\end{IEEEbiography}
\end{document}